\documentclass[runningheads]{llncs}
\usepackage{graphicx}
\usepackage{comment}
\usepackage{amsmath,amssymb} 
\usepackage{color}
\usepackage{subcaption}
\usepackage{mathtools}
\usepackage{paralist} 
\usepackage{colortbl}
\usepackage{nicefrac}
\usepackage{multirow}
\usepackage[table,dvipsnames]{xcolor}
\usepackage{booktabs}
\usepackage{makecell}
\usepackage{gensymb}

\usepackage{afterpage}

\usepackage{pifont}

\usepackage[width=122mm,left=12mm,paperwidth=146mm,height=193mm,top=12mm,paperheight=217mm]{geometry}

\definecolor{yellow}{rgb}{1,1, 0.6}
\definecolor{lightyellow}{rgb}{1,1, 0.8}
\definecolor{orange}{rgb}{1, 0.8, 0.6}
\definecolor{red}{rgb}{1, 0.6, 0.6}

\definecolor{darkyellow}{rgb}{0.8, 0.8, 0.5}
\definecolor{darkred}{rgb}{0.7, 0.3, 0.3}
\definecolor{darkgreen}{rgb}{0.3, 0.7, 0.3}
\definecolor{blue}{rgb}{0, 0, 1.0}
\definecolor{green}{rgb}{0, 1.0, 0}
\definecolor{pink}{rgb}{1, 0.4, 0.7}

\newcommand{\mbf}[1]{{\mathbf{#1}}}

\let\originalleft\left
\let\originalright\right
\renewcommand{\left}{\mathopen{}\mathclose\bgroup\originalleft}
\renewcommand{\right}{\aftergroup\egroup\originalright}

\newcommand{\norm}[1]{\left\lVert#1\right\rVert}

\newcommand{\etal}{\textit{et al}. }
\newcommand{\ie}{\textit{i}.\textit{e}., }

\newcommand{\expo}[1]{\exp\left(#1\right)}

\newcommand{\modeltheta}{\mathrm{\Theta}}
\newcommand{\absrp}{\sigma}

\newcommand{\numsamples}{N}
\newcommand{\numsamplescoarse}{N_c}
\newcommand{\numsamplesfine}{N_f}
\newcommand{\timenear}{t_n}
\newcommand{\timefar}{t_f}
\newcommand{\deltatime}{\delta}

\newcommand{\posxyz}{xyz}
\newcommand{\angletheta}{\theta}
\newcommand{\anglephi}{\phi}
\newcommand{\posall}{\posxyz\angletheta\anglephi}

\newcommand{\numfrequencies}{L}

\newcommand{\ray}{\mathbf{r}}
\newcommand{\Ltrain}{\mathcal{L}}
\newcommand{\raybatch}{\mathcal{R}}
\newcommand{\Ccoarse}{\hat{C}_c(\ray)}
\newcommand{\Cfine}{\hat{C}_f(\ray)}
\newcommand{\Ctrue}{C(\ray)}
\newcommand{\pweight}{w}
\newcommand{\normpweight}{\hat{w}}

\newcommand{\scenename}[1]{\textit{#1}}

\makeatletter
\newcommand{\printfnsymbol}[1]{%
        \textsuperscript{\@fnsymbol{#1}}%
}
\makeatother

\begin{document}
\pagestyle{headings}
\mainmatter
\def\ECCVSubNumber{1473}  

\title{NeRF: Representing Scenes as \\ Neural Radiance Fields for View Synthesis} 


\titlerunning{NeRF: Representing Scenes as Neural Radiance Fields for View Synthesis}
%

\author{Ben Mildenhall$^1$\thanks{Authors contributed equally to this work.}
\quad
Pratul P. Srinivasan$^1$\printfnsymbol{1}
\quad
Matthew Tancik$^1$\printfnsymbol{1} \\
Jonathan T. Barron$^2$
\quad
Ravi Ramamoorthi$^3$
\quad
Ren Ng$^1$ 
}
\institute{$^1$UC Berkeley \quad $^2$Google Research \quad $^3$UC San Diego}
\authorrunning{B. Mildenhall, P. P. Srinivasan, M. Tancik et al.}

\maketitle

\begin{abstract}

We present a method that achieves state-of-the-art results for synthesizing novel views of complex scenes by optimizing an underlying continuous volumetric scene function using a sparse set of input views. Our algorithm represents a scene using a fully-connected (non-convolutional) deep network, whose input is a single continuous 5D coordinate (spatial location $(x,y,z)$ and viewing direction $(\angletheta,\anglephi)$) and whose output is the volume density and view-dependent emitted radiance at that spatial location. We synthesize views by querying 5D coordinates along camera rays and use classic volume rendering techniques to project the output colors and densities into an image. Because volume rendering is naturally differentiable, the only input required to optimize our representation is a set of images with known camera poses. We describe how to effectively optimize neural radiance fields to render photorealistic novel views of scenes with complicated geometry and appearance, and demonstrate results that outperform prior work on neural rendering and view synthesis. View synthesis results are best viewed as videos, so we urge readers to view our supplementary video for convincing comparisons. 

\end{abstract}

\section{Introduction}

In this work, we address the long-standing problem of view synthesis in a new way by directly optimizing parameters of a continuous 5D scene representation to minimize the error of rendering a set of captured images. 

\begin{figure}[t]
\centering
\includegraphics[width=\linewidth]{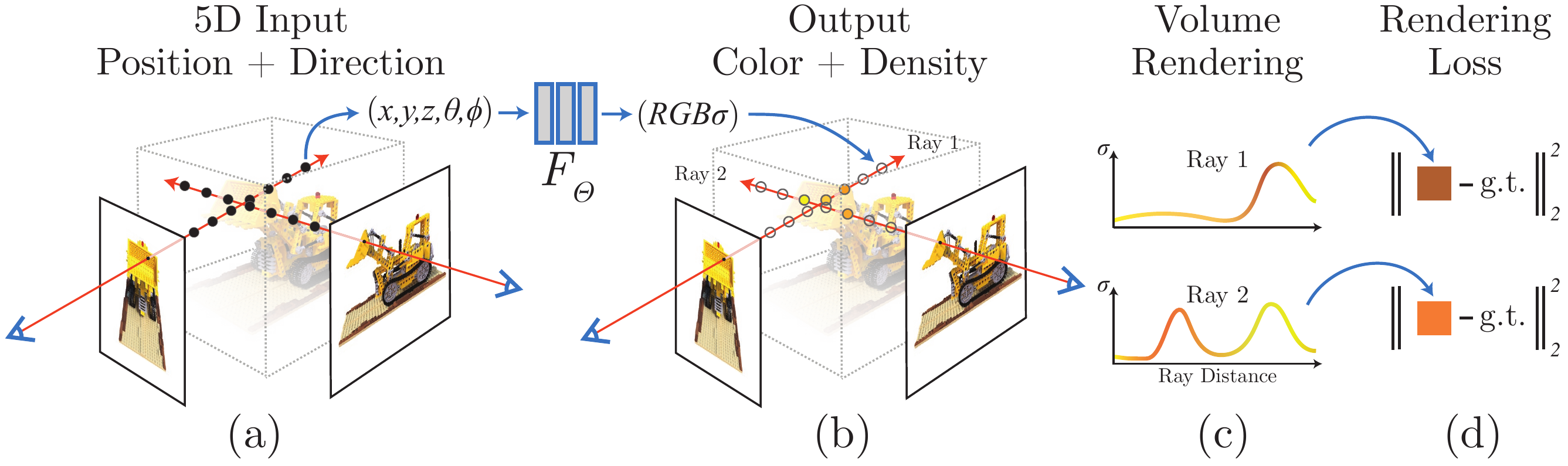}
\caption{
An overview of our neural radiance field scene representation and differentiable rendering procedure. We synthesize images by sampling 5D coordinates (location and viewing direction) along camera rays (a), feeding those locations into an MLP to produce a color and volume density (b), and using volume rendering techniques to composite these values into an image (c). This rendering function is differentiable, so we can optimize our scene representation by minimizing the residual between synthesized and ground truth observed images (d).
}
\label{fig:pipeline}
\end{figure}

We represent a static scene as a continuous 5D function that outputs the radiance emitted in each direction $(\angletheta,\anglephi)$ at each point $(x,y,z)$ in space, and a density at each point which acts like a differential opacity controlling how much radiance is accumulated by a ray passing through $(x,y,z)$.
Our method optimizes a deep fully-connected neural network without any convolutional layers (often referred to as a multilayer perceptron or MLP) to represent this function by regressing from a single 5D coordinate $(x,y,z,\angletheta,\anglephi)$ to a single volume density and view-dependent RGB color. 
To render this \emph{neural radiance field} (NeRF) from a particular viewpoint we:
1) march camera rays through the scene to generate a sampled set of 3D points,
2) use those points and their corresponding 2D viewing directions as input to the neural network to produce an output set of colors and densities, and
3) use classical volume rendering techniques to accumulate those colors and densities into a 2D image.
Because this process is naturally differentiable, we can use gradient descent to optimize this model by minimizing the error between each observed image and the corresponding views rendered from our representation. Minimizing this error across multiple views encourages the network to predict a coherent model of the scene by assigning high volume densities and accurate colors to the locations that contain the true underlying scene content.
Figure~\ref{fig:pipeline} visualizes this overall pipeline.

We find that the basic implementation of optimizing a neural radiance field representation for a complex scene does not converge to a sufficiently high-resolution representation and is inefficient in the required number of samples per camera ray. We address these issues by transforming input 5D coordinates with a positional encoding that enables the MLP to represent higher frequency functions, and we propose a hierarchical sampling procedure to reduce the number of queries required to adequately sample this high-frequency scene representation.

\begin{figure}[t]
\centering
\includegraphics[width=\linewidth]{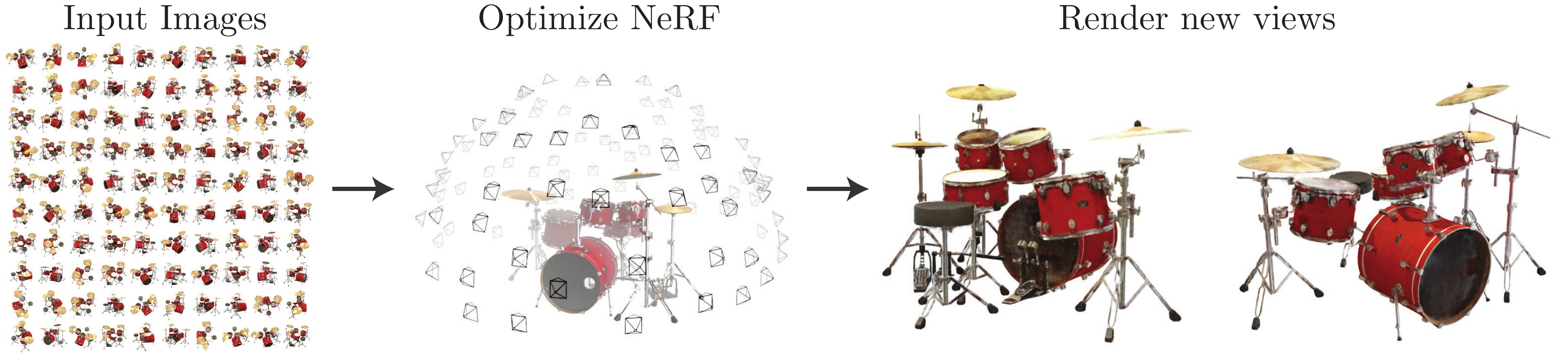}
\caption{We present a method that optimizes a continuous 5D neural radiance field representation (volume density and view-dependent color at any continuous location) of a scene from a set of input images. We use techniques from volume rendering to accumulate samples of this scene representation along rays to render the scene from any viewpoint. Here, we visualize the set of 100 input views of the synthetic \scenename{Drums} scene randomly captured on a surrounding hemisphere, and we show two novel views rendered from our optimized NeRF representation.}
\label{fig:teaser}
\end{figure}

Our approach inherits the benefits of volumetric representations: both can represent complex real-world geometry and appearance and are well suited for gradient-based optimization using projected images. Crucially, our method overcomes the prohibitive storage costs of \emph{discretized} voxel grids when modeling complex scenes at high-resolutions. In summary, our technical contributions are:
\begin{compactenum}[--]
\item An approach for representing continuous scenes with complex geometry and materials as 5D neural radiance fields, parameterized as basic MLP networks.
\item A differentiable rendering procedure based on classical volume rendering techniques, which we use to optimize these representations from standard RGB images. This includes a hierarchical sampling strategy to allocate the MLP's capacity towards space with visible scene content.
\item A positional encoding to map each input 5D coordinate into a higher dimensional space, which enables us to successfully optimize neural radiance fields to represent high-frequency scene content.
\end{compactenum}
We demonstrate that our resulting neural radiance field method quantitatively and qualitatively outperforms state-of-the-art view synthesis methods, including works that fit neural 3D representations to scenes as well as works that train deep convolutional networks to predict sampled volumetric representations. As far as we know, this paper presents the first continuous neural scene representation that is able to render high-resolution photorealistic novel views of real objects and scenes from RGB images captured in natural settings.

\section{Related Work}

A promising recent direction in computer vision is encoding objects and scenes in the weights of an MLP that directly maps from a 3D spatial location to an implicit representation of the shape, such as the signed distance~\cite{sdf} at that location. However, these methods have so far been unable to reproduce realistic scenes with complex geometry with the same fidelity as techniques that represent scenes using discrete representations such as triangle meshes or voxel grids. In this section, we review these two lines of work and contrast them with our approach, which enhances the capabilities of neural scene representations to produce state-of-the-art results for rendering complex realistic scenes. 

A similar approach of using MLPs to map from low-dimensional coordinates to colors has also been used for representing other graphics functions such as images~\cite{stanley2007compositional}, textured materials~\cite{henzler20,texturefields,rainer20,rainer19}, and indirect illumination values~\cite{rrf}.

\paragraph{\textbf{Neural 3D shape representations}}

Recent work has investigated the implicit representation of continuous 3D shapes as level sets by optimizing deep networks that map $\posxyz$ coordinates to signed distance functions~\cite{jiang2020,deepsdf} or occupancy fields~\cite{genova2020,occupancynet}. However, these models are limited by their requirement of access to ground truth 3D geometry, typically obtained from synthetic 3D shape datasets such as ShapeNet~\cite{shapenet}. 
Subsequent work has relaxed this requirement of ground truth 3D shapes by formulating differentiable rendering functions that allow neural implicit shape representations to be optimized using only 2D images. Niemeyer \etal~\cite{diffvolumetric} represent surfaces as 3D occupancy fields and use a numerical method to find the surface intersection for each ray, then calculate an exact derivative using implicit differentiation. Each ray intersection location is provided as the input to a neural 3D texture field that predicts a diffuse color for that point. Sitzmann \etal~\cite{srn} use a less direct neural 3D representation that simply outputs a feature vector and RGB color at each continuous 3D coordinate, and propose a differentiable rendering function consisting of a recurrent neural network that marches along each ray to decide where the surface is located.

Though these techniques can potentially represent complicated and high-resolution geometry, they have so far been limited to simple shapes with low geometric complexity, resulting in oversmoothed renderings. We show that an alternate strategy of optimizing networks to encode 5D radiance fields (3D volumes with 2D view-dependent appearance) can represent higher-resolution geometry and appearance to render photorealistic novel views of complex scenes.

\paragraph{\textbf{View synthesis and image-based rendering}}

Given a dense sampling of views, photorealistic novel views can be reconstructed by simple light field sample interpolation techniques~\cite{levoy96,cohen96,davis12}. For novel view synthesis with sparser view sampling, the computer vision and graphics communities have made significant progress by predicting traditional geometry and appearance representations from observed images.
One popular class of approaches uses mesh-based representations of scenes with either diffuse~\cite{waechter14} or view-dependent~\cite{buehler01,debevec96,wood00} appearance.
Differentiable rasterizers~\cite{dibr,genova18,softras,opendr} or pathtracers~\cite{redner,Mitsuba2} can directly optimize mesh representations to reproduce a set of input images using gradient descent.
However, gradient-based mesh optimization based on image reprojection is often difficult, likely because of local minima or poor conditioning of the loss landscape. Furthermore, this strategy requires a template mesh with fixed topology to be provided as an initialization before optimization~\cite{redner}, which is typically unavailable for unconstrained real-world scenes.

Another class of methods use volumetric representations to address the task of high-quality photorealistic view synthesis from a set of input RGB images.
Volumetric approaches are able to realistically represent complex shapes and materials, are well-suited for gradient-based optimization, and tend to produce less visually distracting artifacts than mesh-based methods.
Early volumetric approaches used observed images to directly color voxel grids~\cite{kutulakos00,seitz99,szeliski98}. More recently, several methods~\cite{flynn19,henzler18,kar17,mildenhall19,penner17,srinivasan19,tulsiani17,zhou18} have used large datasets of multiple scenes to train deep networks that predict a sampled volumetric representation from a set of input images, and then use either alpha-compositing~\cite{porter84} or learned compositing along rays to render novel views at test time.
Other works have optimized a combination of convolutional networks (CNNs) and sampled voxel grids for each specific scene, such that the CNN can compensate for discretization artifacts from low resolution voxel grids~\cite{deepvoxels} or allow the predicted voxel grids to vary based on input time or animation controls~\cite{neuralvolumes}.
While these volumetric techniques have achieved impressive results for novel view synthesis, their ability to scale to higher resolution imagery is fundamentally limited by poor time and space complexity due to their discrete sampling --- rendering higher resolution images requires a finer sampling of 3D space.
We circumvent this problem by instead encoding a \emph{continuous} volume within the parameters of a deep fully-connected neural network, which not only produces significantly higher quality renderings than prior volumetric approaches, but also requires just a fraction of the storage cost of those \emph{sampled} volumetric representations.

\section{Neural Radiance Field Scene Representation}

We represent a continuous scene as a 5D vector-valued function whose input is a 3D location $\mathbf{x} = (x,y,z)$ and 2D viewing direction $(\theta, \phi)$, and whose output is an emitted color $\mathbf{c} = (r,g,b)$ and volume density $\sigma$. In practice, we express direction as a 3D Cartesian unit vector $\mathbf{d}$. We approximate this continuous 5D scene representation with an MLP network $F_{\modeltheta} : (\mathbf{x}, \mathbf{d}) \to (\mathbf{c},\sigma)$ and optimize its weights $\modeltheta$ to map from each input 5D coordinate to its corresponding volume density and directional emitted color. 

We encourage the representation to be multiview consistent by restricting the network to predict the volume density $\sigma$ as a function of only the location $\mathbf{x}$, while allowing the RGB color $\mathbf{c}$ to be predicted as a function of both location and viewing direction. To accomplish this, the MLP $F_\modeltheta$ first processes the input 3D coordinate $\mathbf{x}$ with 8 fully-connected layers (using ReLU activations and 256 channels per layer), and outputs $\sigma$ and a 256-dimensional feature vector. This feature vector is then concatenated with the camera ray's viewing direction and passed to one additional fully-connected layer (using a ReLU activation and 128 channels) that output the view-dependent RGB color.

\begin{figure}[t]
\centering
\includegraphics[width=\linewidth]{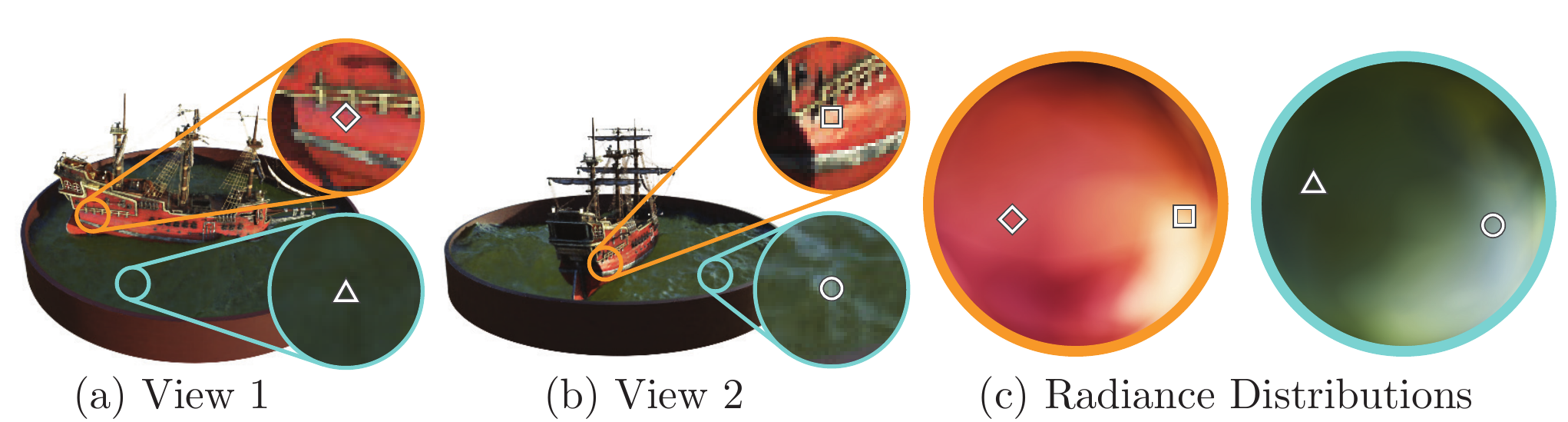}
\caption{A visualization of view-dependent emitted radiance. Our neural radiance field representation outputs RGB color as a 5D function of both spatial position $\mathbf x$ and viewing direction $\mathbf d$. Here, we visualize example directional color distributions for two spatial locations in our neural representation of the \scenename{Ship} scene. 
In (a) and (b), we show the appearance of two fixed 3D points from two different camera positions: one on the side of the ship (orange insets) and one on the surface of the water (blue insets).
Our method predicts the changing specular appearance of these two 3D points, and in (c) we show how this behavior generalizes continuously across the whole hemisphere of viewing directions.
}
\label{fig:viewdirs}
\end{figure}

See Fig.~\ref{fig:viewdirs} for an example of how our method uses the input viewing direction to represent non-Lambertian effects.
As shown in Fig.~\ref{fig:ablations}, a model trained without view dependence (only $\mathbf x$ as input) has difficulty representing specularities.

\section{Volume Rendering with Radiance Fields}
\label{sec:rendering}

Our 5D neural radiance field represents a scene as the volume density and directional emitted radiance at any point in space. We render the color of any ray passing through the scene using principles from classical volume rendering~\cite{kajiya84}. The volume density $\sigma(\mathbf{x})$ can be interpreted as the differential probability of a ray terminating at an infinitesimal particle at location $\mathbf{x}$. The expected color $C(\mathbf{r})$ of camera ray $\mathbf{r}(t)=\mathbf{o} + t\mathbf{d}$ with near and far bounds $\timenear$ and $\timefar$ is: 
\begin{equation}
\Ctrue = \int_{\timenear}^{\timefar}T(t)\absrp(\mathbf{r}(t))\mathbf{c}(\mathbf{r}(t),\mathbf{d})dt\,, \textrm{ where }
T(t) = \expo{-\int_{\timenear}^{t}\absrp(\mathbf{r}(s))ds}\,.
\end{equation}
The function $T(t)$ denotes the accumulated transmittance along the ray from $\timenear$ to $t$, \ie the probability that the ray travels from $\timenear$ to $t$ without hitting any other particle. Rendering a view from our continuous neural radiance field requires estimating this integral $\Ctrue$ for a camera ray traced through each pixel of the desired virtual camera.

We numerically estimate this continuous integral using quadrature. Deterministic quadrature, which is typically used for rendering discretized voxel grids, would effectively limit our representation's resolution because the MLP would only be queried at a fixed discrete set of locations. Instead, we use a stratified sampling approach where we partition $[\timenear, \timefar]$ into $\numsamples$ evenly-spaced bins and then draw one sample uniformly at random from within each bin:
\begin{equation}
    t_i \sim \mathcal{U} \left[ \timenear + \frac{i-1}{\numsamples}(\timefar-\timenear),\,\, \timenear + \frac{i}{\numsamples}(\timefar-\timenear) \right]\,.
    \label{eq:stratified}
\end{equation} 
Although we use a discrete set of samples to estimate the integral, stratified sampling enables us to represent a continuous scene representation because it results in the MLP being evaluated at continuous positions over the course of optimization. We use these samples to estimate $\Ctrue$ with the quadrature rule discussed in the volume rendering review by Max~\cite{max95}:
\begin{equation}
\label{eqn:render_coarse}
\hat{C}(\mathbf{r})=\sum_{i=1}^{\numsamples}T_i (1-\expo{-\absrp_i \deltatime_i}) \mathbf{c}_i\,, \textrm{ where }
T_i=\expo{- \sum_{j=1}^{i-1} \absrp_j \deltatime_j}\,,
\end{equation}
where $\deltatime_i = t_{i+1} - t_i $ is the distance between adjacent samples. 
This function for calculating $\hat{C}(\mathbf{r})$ from the set of $(\mathbf{c}_i, \absrp_i)$ values is trivially differentiable and reduces to traditional alpha compositing with alpha values $\alpha_i = 1-\expo{-\absrp_i \deltatime_i}$.

\section{Optimizing a Neural Radiance Field}

In the previous section we have described the core components necessary for modeling a scene as a neural radiance field and rendering novel views from this representation. However, we observe that these components are not sufficient for achieving state-of-the-art quality, as demonstrated in Section~\ref{sec:ablations}). We introduce two improvements to enable representing high-resolution complex scenes. The first is a positional encoding of the input coordinates that assists the MLP in representing high-frequency functions, and the second is a hierarchical sampling procedure that allows us to efficiently sample this high-frequency representation.

\newcommand{\resultscropwidth}{1.14in}

\newcommand{\cropablationB}[1]{
  \makecell{
  \includegraphics[trim={366px 288px 210px 288px}, clip, width=\resultscropwidth]{#1} 
  }
}

\begin{figure}[t]
\centering
\scriptsize
\begin{tabular}{@{}c@{}c@{}c@{}c@{}}
\cropablationB{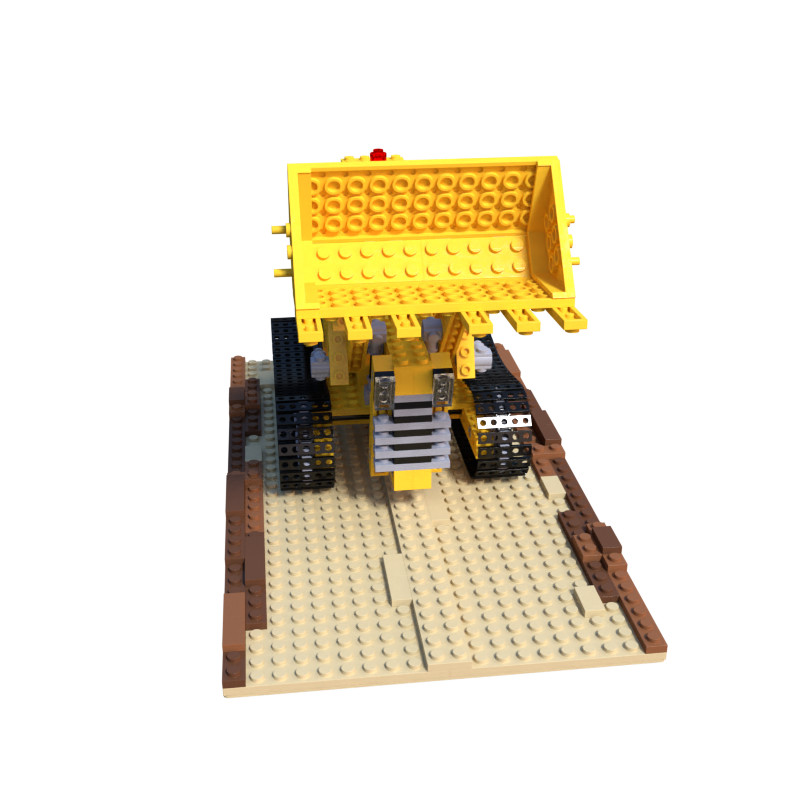} &
\cropablationB{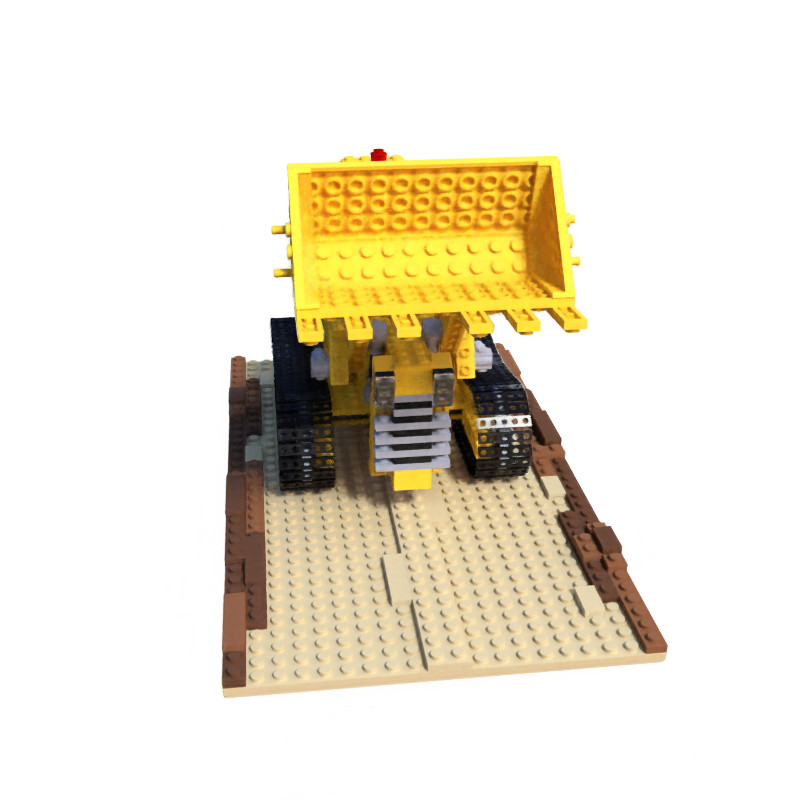} &
\cropablationB{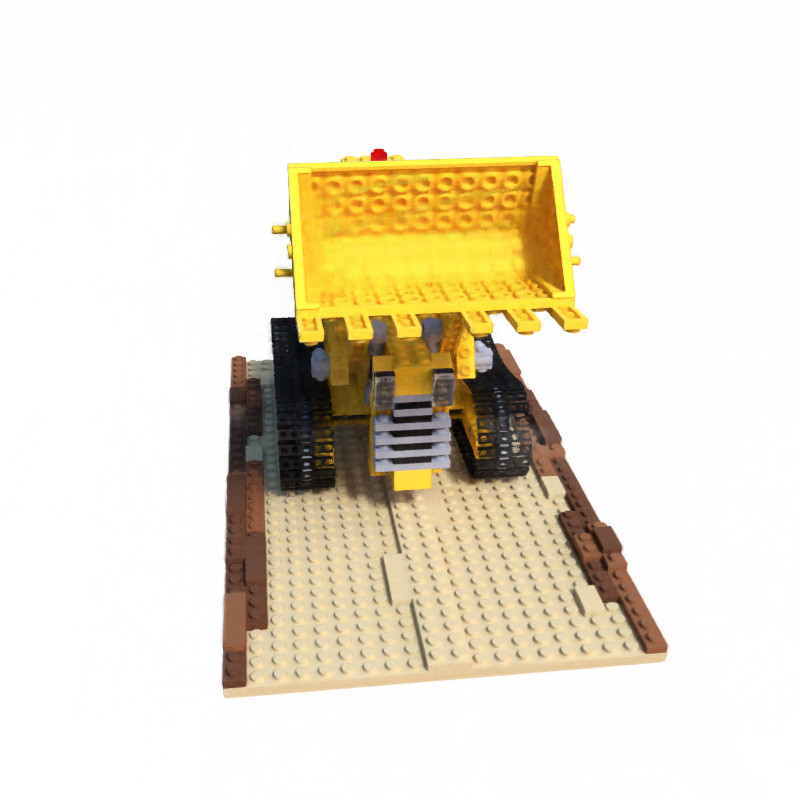} &
\cropablationB{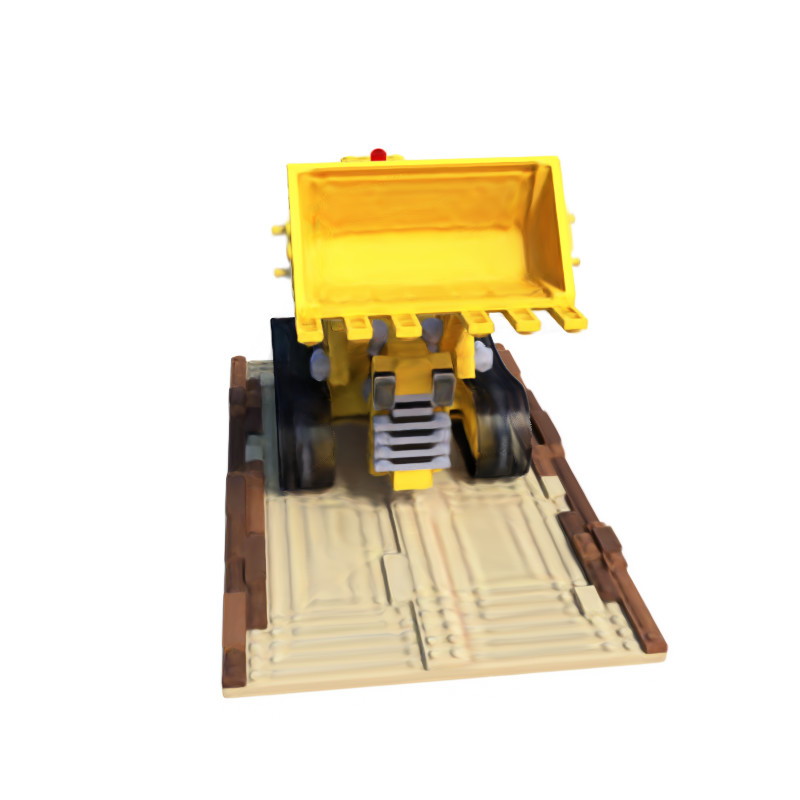} \\
Ground Truth & Complete Model & No View Dependence & No Positional Encoding
\end{tabular}
\caption{Here we visualize how our full model benefits from representing view-dependent emitted radiance and from passing our input coordinates through a high-frequency positional encoding. Removing view dependence prevents the model from recreating the specular reflection on the bulldozer tread. Removing the positional encoding drastically decreases the model's ability to represent high frequency geometry and texture, resulting in an oversmoothed appearance.}
\label{fig:ablations}
\end{figure}

\subsection{Positional encoding}
\label{sec:posenc}

Despite the fact that neural networks are universal function approximators~\cite{universal}, we found that having the network $F_\modeltheta$ directly operate on $\posall$ input coordinates results in renderings that perform poorly at representing high-frequency variation in color and geometry. This is consistent with recent work by Rahaman \etal~\cite{rahaman2018}, which shows that deep networks are biased towards learning lower frequency functions. They additionally show that mapping the inputs to a higher dimensional space using high frequency functions before passing them to the network enables better fitting of data that contains high frequency variation.

\newcommand{\dummypos}{p}

We leverage these findings in the context of neural scene representations, and show that reformulating $F_\modeltheta$ as a composition of two functions ${F_\modeltheta = F'_\modeltheta \circ \gamma }$, one learned and one not, significantly improves performance (see Fig.~\ref{fig:ablations} and Table~\ref{table:ablations}).
Here $\gamma$ is a mapping from $\mathbb{R}$ into a higher dimensional space $\mathbb{R}^{2\numfrequencies}$, and $F'_\modeltheta$ is still simply a regular MLP. Formally, the encoding function we use is: 
\begin{equation}
    \gamma(\dummypos) = \left(
    \begin{array}{ccccc}
    \sin\left(2^0 \pi \dummypos\right), &
    \cos\left(2^0 \pi \dummypos\right), &
    \cdots, &
    \sin\left(2^{\numfrequencies-1} \pi \dummypos\right), &
    \cos\left(2^{\numfrequencies-1} \pi \dummypos\right)
    \end{array} \right) \,.
\label{eq:enc}
\end{equation}
This function $\gamma(\cdot)$ is applied separately to each of the three coordinate values in $\mathbf{x}$ (which are normalized to lie in $[-1, 1]$) and to the three components of the Cartesian viewing direction unit vector $\mathbf{d}$ (which by construction lie in $[-1, 1]$). 
In our experiments, we set $L=10$ for $\gamma(\mathbf x)$ and $L=4$ for $\gamma(\mathbf d)$. 

A similar mapping is used in the popular Transformer architecture~\cite{transformer}, where it is referred to as a \emph{positional encoding}. However, Transformers use it for a different goal of providing the discrete positions of tokens in a sequence as input to an architecture that does not contain any notion of order. In contrast, we use these functions to map continuous input coordinates into a higher dimensional space to enable our MLP to more easily approximate a higher frequency function. Concurrent work on a related problem of modeling 3D protein structure from projections~\cite{cryoem} also utilizes a similar input coordinate mapping.

\subsection{Hierarchical volume sampling}
\label{sec:hierarchical}

Our rendering strategy of densely evaluating the neural radiance field network at $\numsamples$ query points along each camera ray is inefficient: free space and occluded regions that do not contribute to the rendered image are still sampled repeatedly.
We draw inspiration from early work in volume rendering~\cite{levoy90} and propose a hierarchical representation that increases rendering efficiency by allocating samples proportionally to their expected effect on the final rendering.

Instead of just using a single network to represent the scene, we simultaneously optimize two networks: one ``coarse'' and one ``fine''.
We first sample a set of $\numsamplescoarse$ locations using stratified sampling, and evaluate the ``coarse'' network at these locations as described in Eqns.~\ref{eq:stratified} and \ref{eqn:render_coarse}. Given the output of this ``coarse'' network, we then produce a more informed sampling of points along each ray where samples are biased towards the relevant parts of the volume. To do this, we first rewrite the alpha composited color from the coarse network $\Ccoarse$ in Eqn.~\ref{eqn:render_coarse} as a weighted sum of all sampled colors $c_i$ along the ray:
\begin{equation}
  \Ccoarse=\sum_{i=1}^{\numsamplescoarse}\pweight_i c_i\,, \quad\,\,
  \pweight_i = T_i(1-\expo{-\absrp_i \deltatime_i})\,.
  \label{eqn:weights}
\end{equation}
Normalizing these weights as $\normpweight_i = \nicefrac{\pweight_i}{\sum_{j=1}^{\numsamplescoarse} \pweight_j}$ produces a piecewise-constant PDF along the ray.
We sample a second set of $\numsamplesfine$ locations from this distribution using inverse transform sampling, evaluate our ``fine'' network at the union of the first and second set of samples, and compute the final rendered color of the ray $\Cfine$ using Eqn.~\ref{eqn:render_coarse} but using all $\numsamplescoarse+\numsamplesfine$ samples. This procedure allocates more samples to regions we expect to contain visible content. This addresses a similar goal as importance sampling, but we use the sampled values as a nonuniform discretization of the whole integration domain rather than treating each sample as an independent probabilistic estimate of the entire integral.

\subsection{Implementation details}

We optimize a separate neural continuous volume representation network for each scene. This requires only a dataset of captured RGB images of the scene, the corresponding camera poses and intrinsic parameters, and scene bounds (we use ground truth camera poses, intrinsics, and bounds for synthetic data, and use the COLMAP structure-from-motion package~\cite{colmap} to estimate these parameters for real data). At each optimization iteration, we randomly sample a batch of camera rays from the set of all pixels in the dataset, and then follow the hierarchical sampling described in Sec.~\ref{sec:hierarchical} to query $\numsamplescoarse$ samples from the coarse network and $\numsamplescoarse+\numsamplesfine$ samples from the fine network. We then use the volume rendering procedure described in Sec.~\ref{sec:rendering} to render the color of each ray from both sets of samples. Our loss is simply the total squared error between the rendered and true pixel colors for both the coarse and fine renderings:
\begin{equation}
\Ltrain=\sum_{\ray\in\raybatch}\left[\norm{\Ccoarse-\Ctrue}_2^2+\norm{\Cfine-\Ctrue}_2^2\right]
\end{equation}
where $\raybatch$ is the set of rays in each batch, and $\Ctrue$, $\Ccoarse$, and $\Cfine$ are the ground truth, coarse volume predicted, and fine volume predicted RGB colors for ray $\ray$ respectively. Note that even though the final rendering comes from $\Cfine$, we also minimize the loss of $\Ccoarse$ so that the weight distribution from the coarse network can be used to allocate samples in the fine network.

In our experiments, we use a batch size of 4096 rays, each sampled at $\numsamplescoarse=64$ coordinates in the coarse volume and $\numsamplesfine=128$ additional coordinates in the fine volume. We use the Adam optimizer~\cite{KingmaB15} with a learning rate that begins at $5 \times 10^{-4}$ and decays exponentially to $5 \times 10^{-5}$ over the course of optimization (other Adam hyperparameters are left at default values of $\beta_1=0.9$, $\beta_2=0.999$, and $\epsilon=10^{-7}$). The optimization for a single scene typically take around 100--300k iterations to converge on a single NVIDIA V100 GPU (about 1--2 days).

\section{Results}

We quantitatively (Tables~\ref{table:results}) and qualitatively (Figs.~\ref{fig:synthresults} and~\ref{fig:realresults}) show that our method outperforms prior work, and provide extensive ablation studies to validate our design choices (Table~\ref{table:ablations}). We urge the reader to view our supplementary video to better appreciate our method's significant improvement over baseline methods when rendering smooth paths of novel views.

\subsection{Datasets}

\paragraph{\textbf{Synthetic renderings of objects}}
\label{subsec:goodsynth}

We first show experimental results on two datasets of synthetic renderings of objects (Table~\ref{table:results}, ``Diffuse Synthetic $360\degree$'' and ``Realistic Synthetic $360\degree$''). The DeepVoxels~\cite{deepvoxels} dataset contains four Lambertian objects with simple geometry. Each object is rendered at $512\times 512$ pixels from viewpoints sampled on the upper hemisphere (479 as input and 1000 for testing). We additionally generate our own dataset containing pathtraced images of eight objects that exhibit complicated geometry and realistic non-Lambertian materials. Six are rendered from viewpoints sampled on the upper hemisphere, and two are rendered from viewpoints sampled on a full sphere. We render 100 views of each scene as input and 200 for testing, all at $800\times 800$ pixels.

\newcommand{\resultsfigwidth}{1.14in}
\renewcommand{\resultscropwidth}{0.69in}

\newcommand{\cropship}[1]{
  \makecell{
  \includegraphics[trim={331px 438px 325px 218px}, clip, width=\resultscropwidth]{#1} \\
  \includegraphics[trim={352px 298px 320px 374px}, clip, width=\resultscropwidth]{#1}
  }
}

\newcommand{\croplego}[1]{
  \makecell{
  \includegraphics[trim={440px 420px 218px 238px}, clip, width=\resultscropwidth]{#1} \\
  \includegraphics[trim={541px 289px 91px 343px}, clip, width=\resultscropwidth]{#1}
  }
}

\newcommand{\cropmic}[1]{
  \makecell{
  \includegraphics[trim={352px 261px 274px 365px}, clip, width=\resultscropwidth]{#1} \\
  \includegraphics[trim={182px 583px 504px 103px}, clip, width=\resultscropwidth]{#1}
  }
}

\newcommand{\cropmat}[1]{
  \makecell{
  \includegraphics[trim={194px 274px 402px 322px}, clip, width=\resultscropwidth]{#1} \\
  \includegraphics[trim={418px 363px 258px 313px}, clip, width=\resultscropwidth]{#1}
  }
}

\begin{figure}[t]
\centering
\scriptsize
\begin{tabular}{@{}c@{}c@{}c@{}c@{}c@{}c@{}}
\makecell[c]{
\includegraphics[trim={0px 0px 0px 100px}, clip, width=\resultsfigwidth]{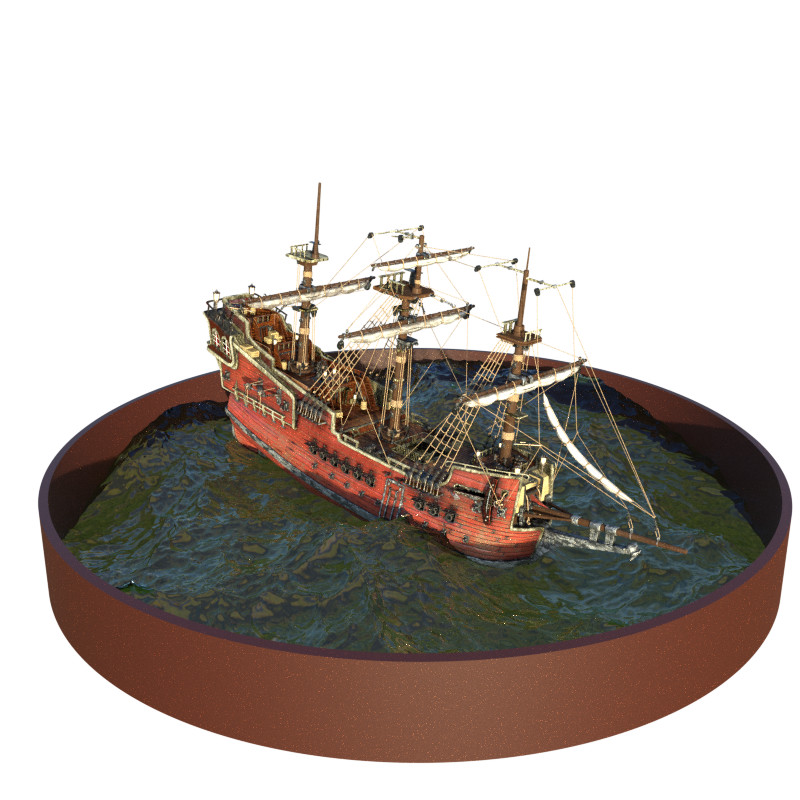}
\\
\scenename{Ship}
}
&
\cropship{figs/synth_results/ship/gt.jpg} &
\cropship{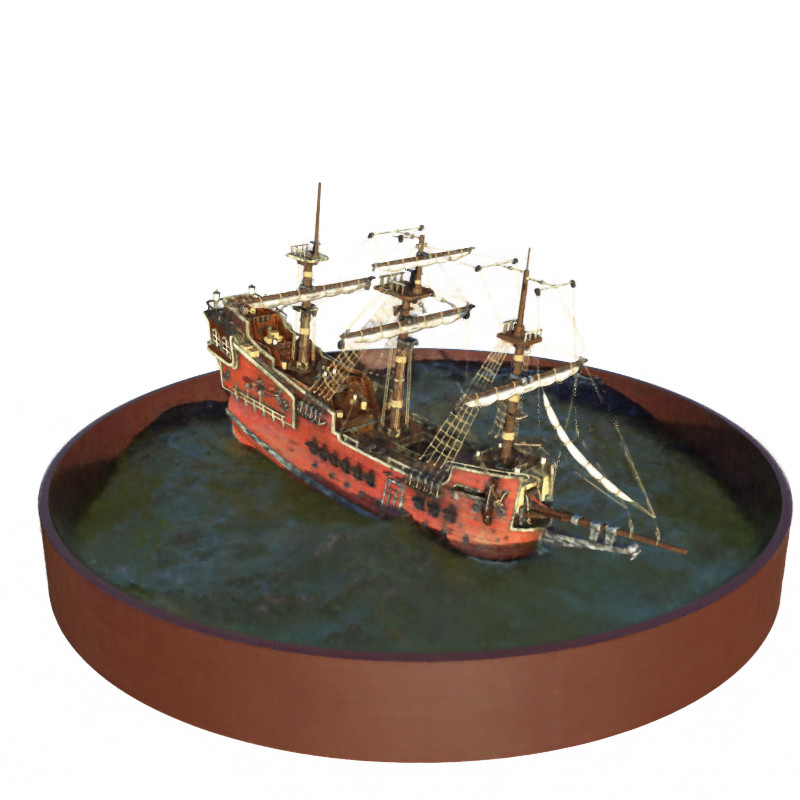} &
\cropship{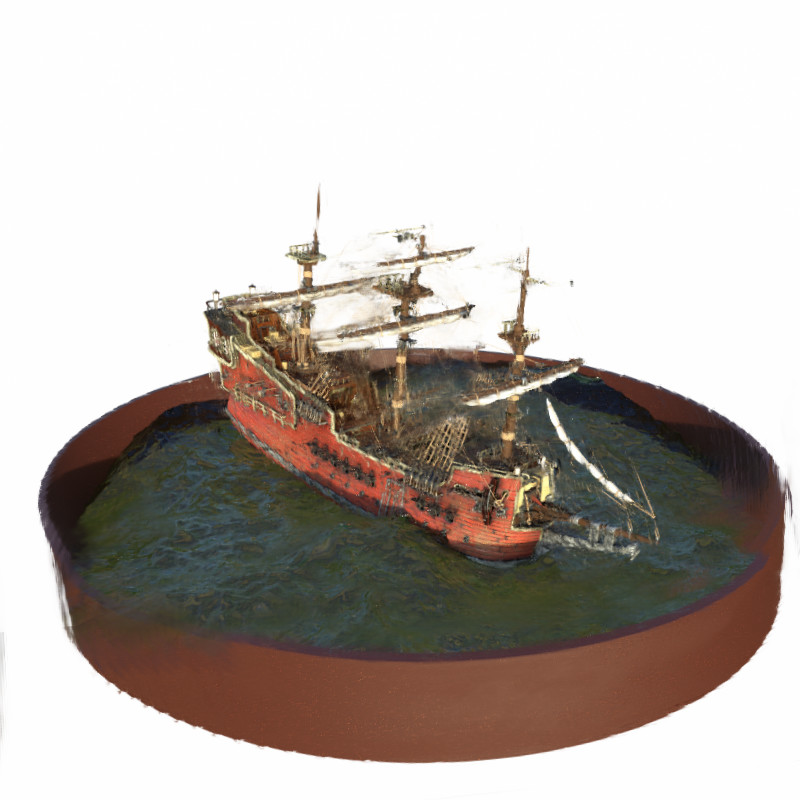} &
\cropship{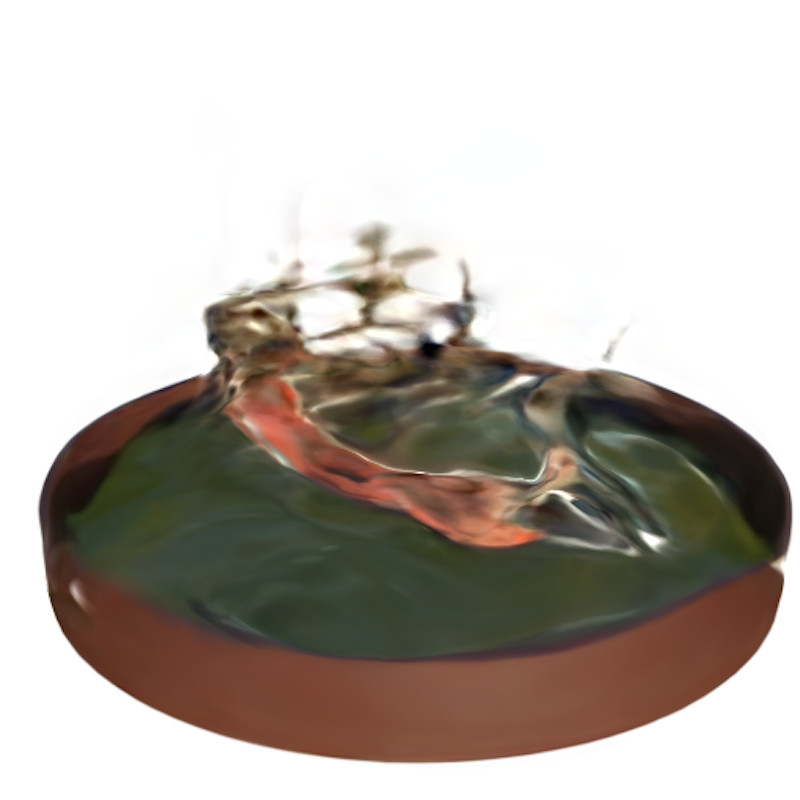} &
\cropship{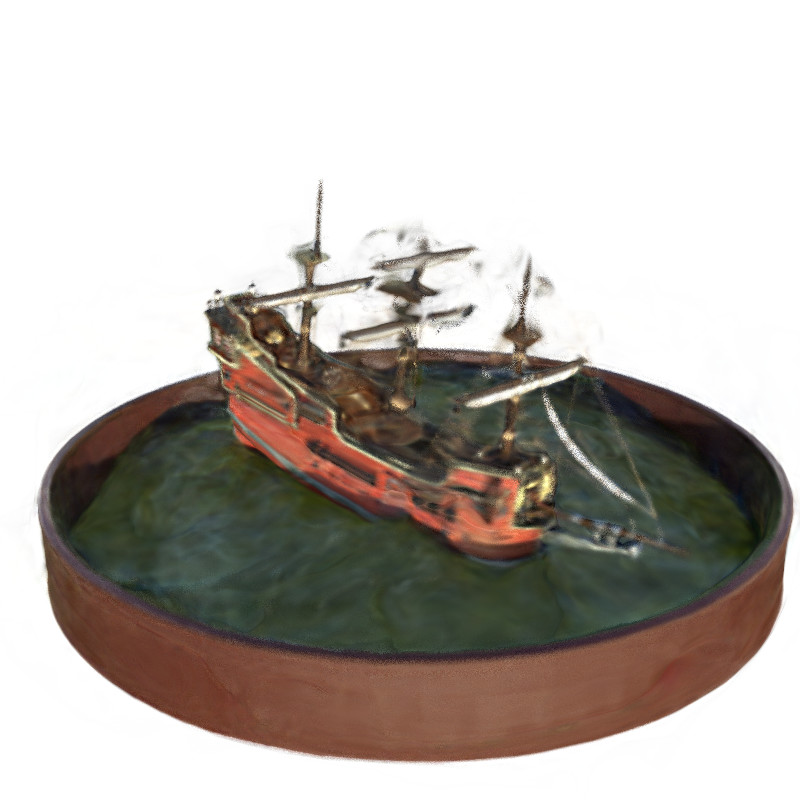} \\
\makecell[c]{
\includegraphics[trim={86px 178px 18px 100px}, clip, width=\resultsfigwidth]{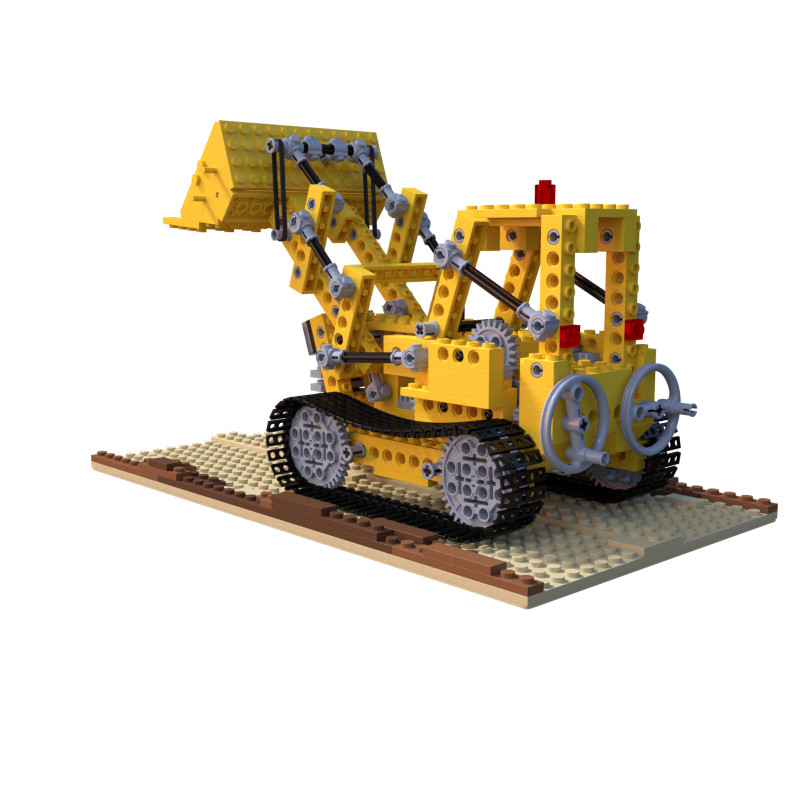}
\\
\scenename{Lego}
}
&
\croplego{figs/synth_results/lego/gt.jpg} &
\croplego{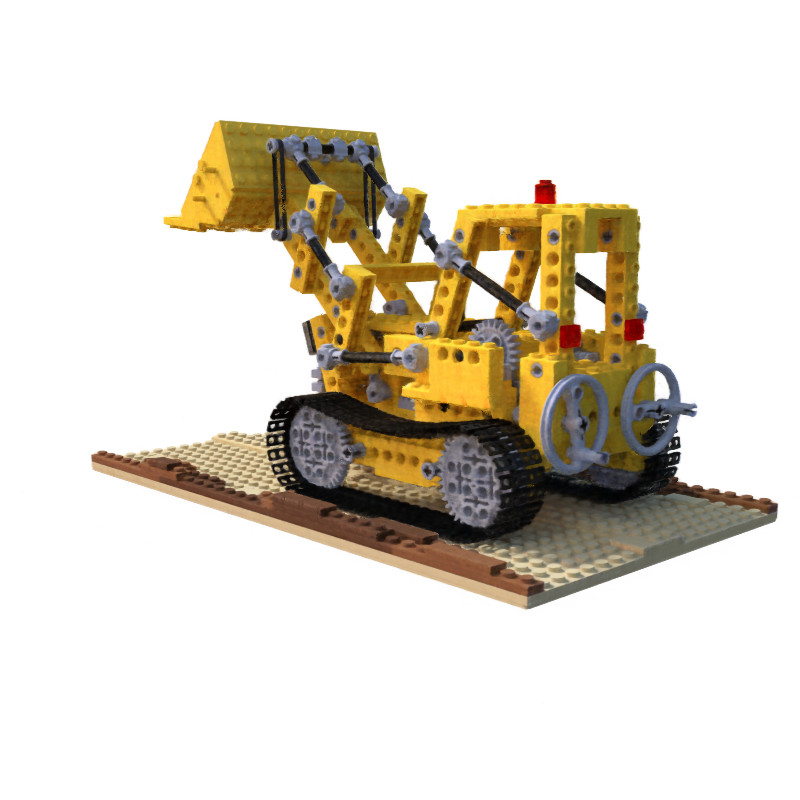} &
\croplego{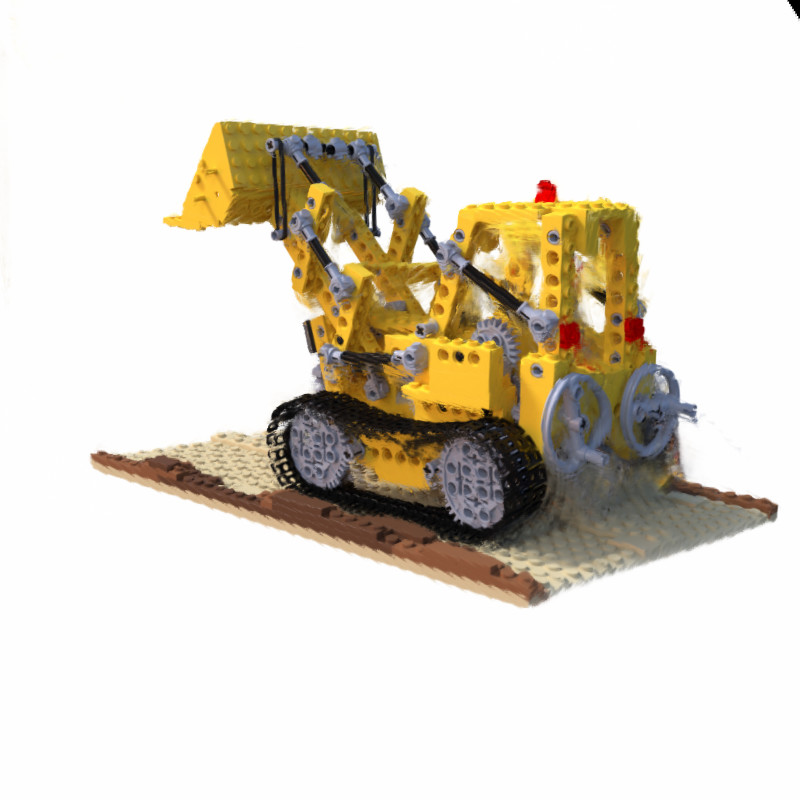} &
\croplego{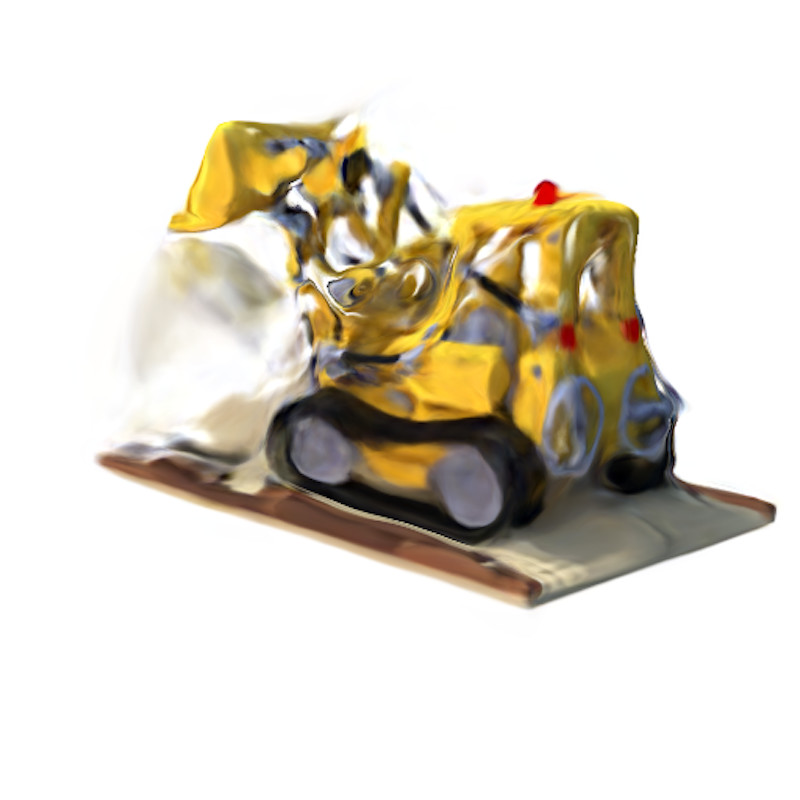} &
\croplego{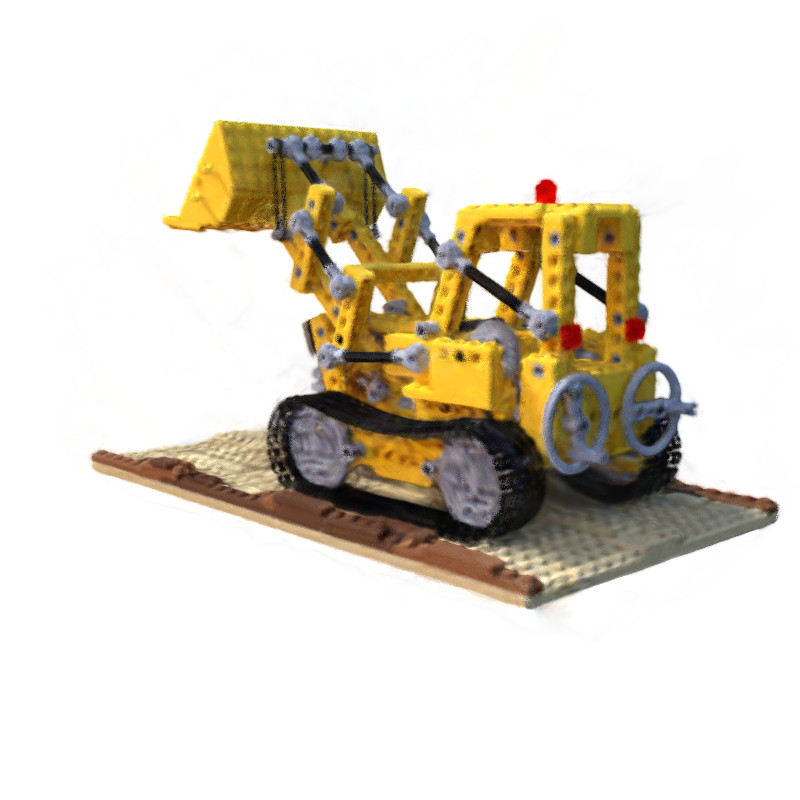} \\
\makecell[c]{
\includegraphics[trim={95px 95px 65px 50px}, clip, width=\resultsfigwidth]{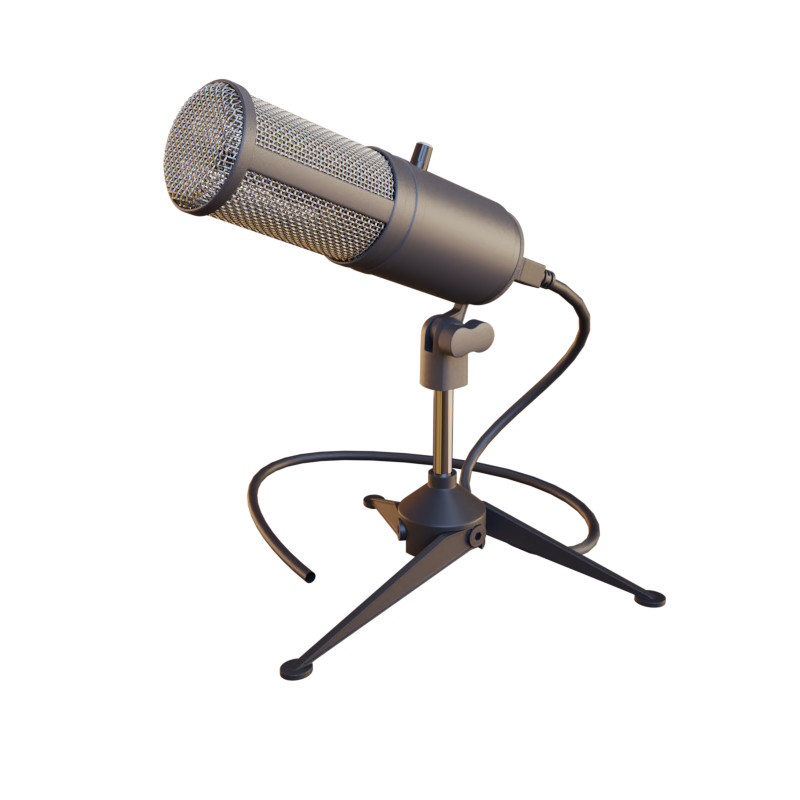} 
\\
\scenename{Microphone}
}
&
\cropmic{figs/synth_results/mic/gt.jpg} &
\cropmic{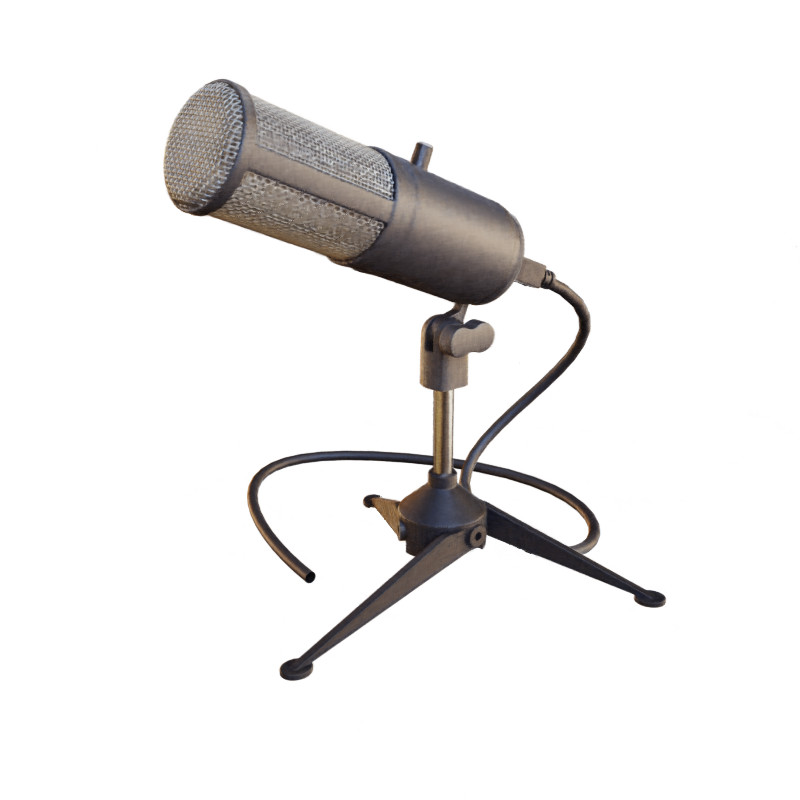} &
\cropmic{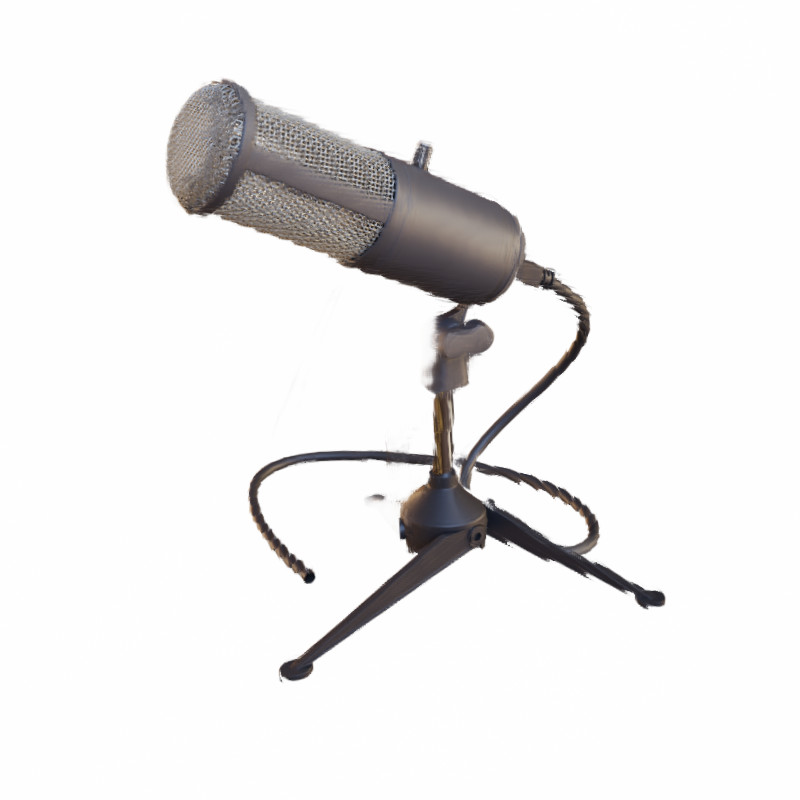} &
\cropmic{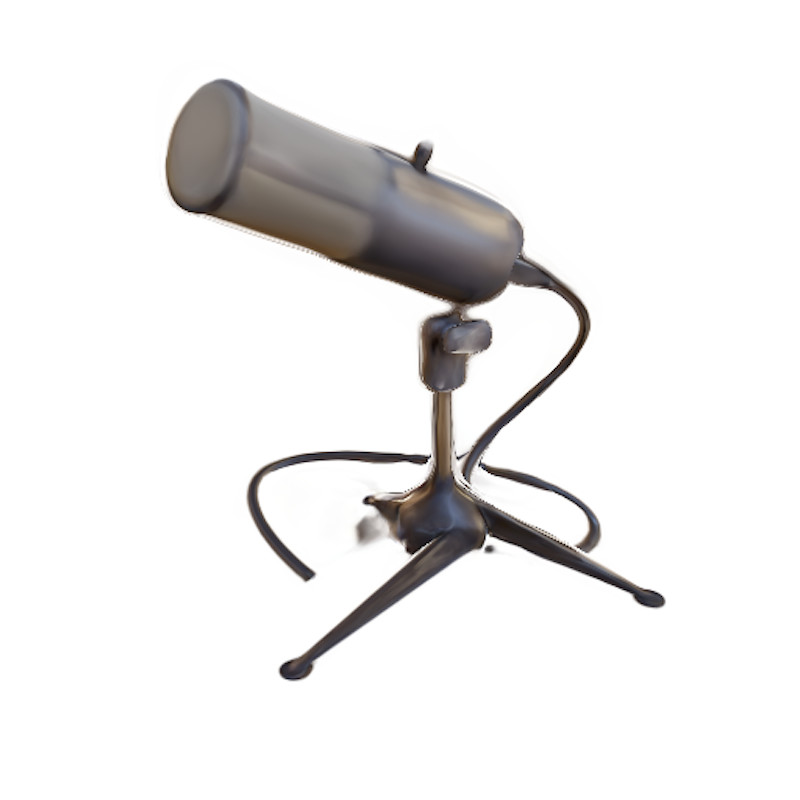} &
\cropmic{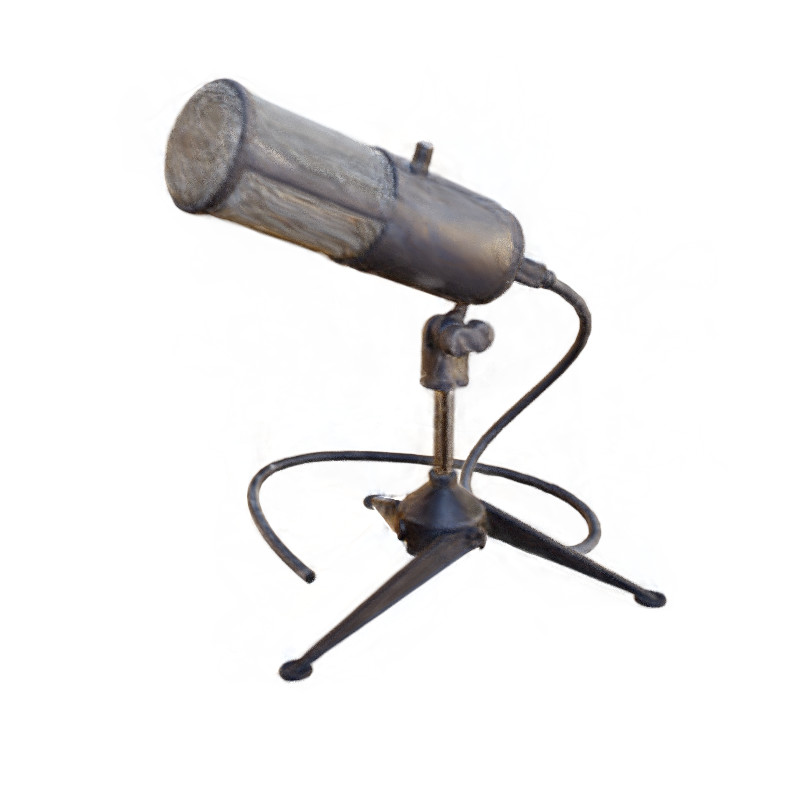} \\
\makecell[c]{
\includegraphics[trim={0px 0px 0px 0px}, clip, width=\resultsfigwidth]{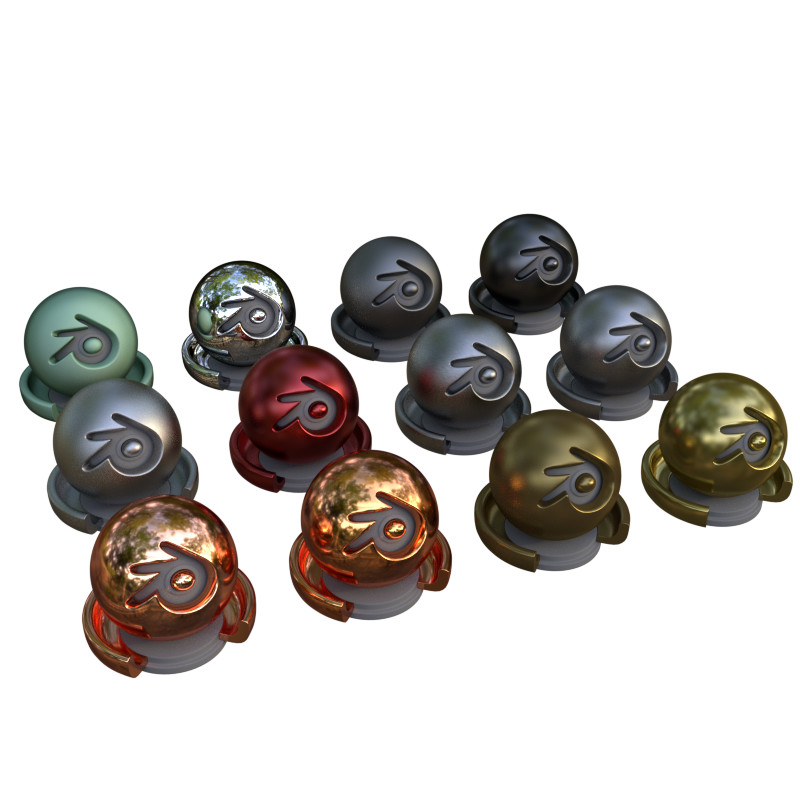} 
\\
\scenename{Materials}
}
&
\cropmat{figs/synth_results/materials/gt.jpg} &
\cropmat{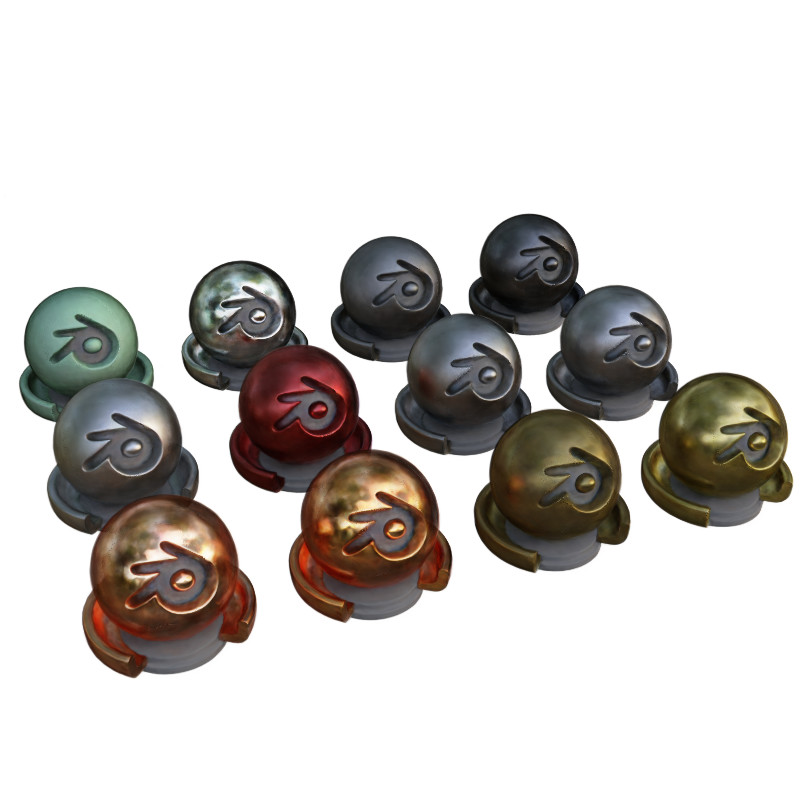} &
\cropmat{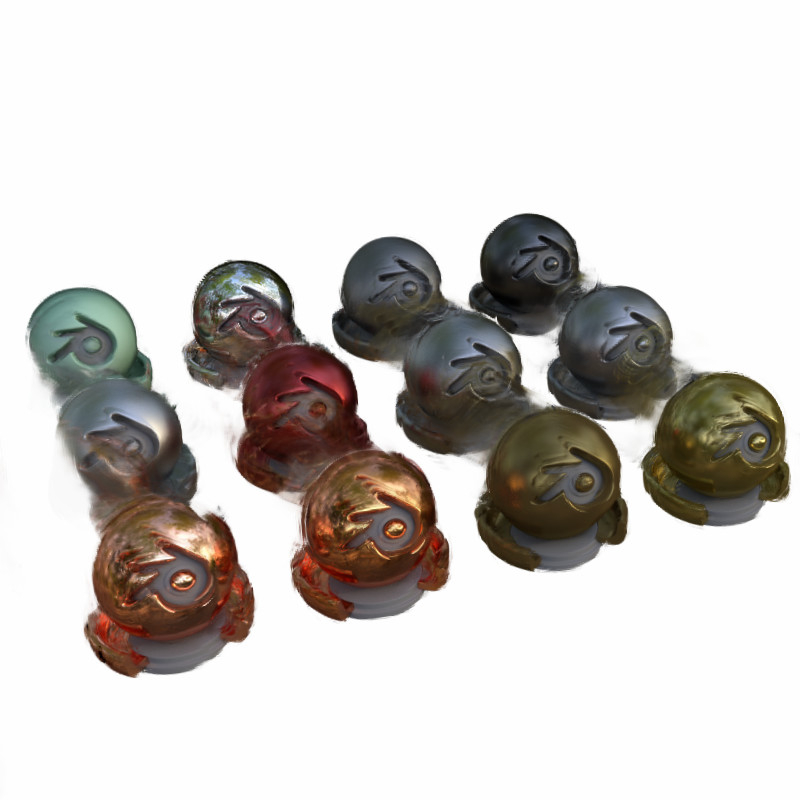} &
\cropmat{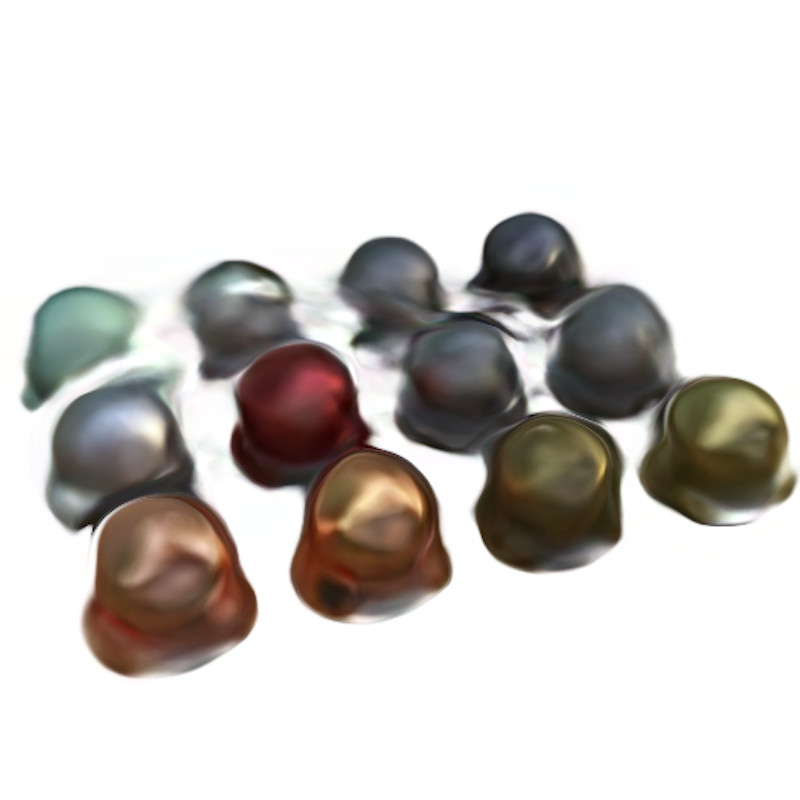} &
\cropmat{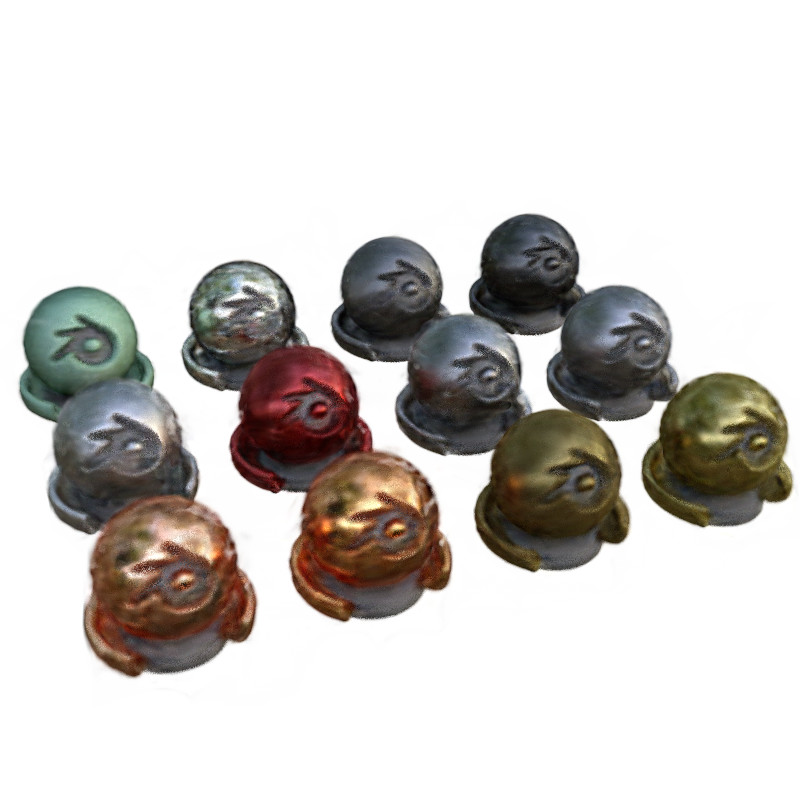} \\
& Ground Truth & NeRF (ours) & LLFF~\cite{mildenhall19} & SRN~\cite{srn} & NV~\cite{neuralvolumes}
\end{tabular} 
\caption{Comparisons on test-set views for scenes from our new synthetic dataset generated with a physically-based renderer. Our method is able to recover fine details in both geometry and appearance, such as \scenename{Ship}'s rigging, \scenename{Lego}'s gear and treads, \scenename{Microphone}'s shiny stand and mesh grille, and \scenename{Material}'s non-Lambertian reflectance. LLFF exhibits banding artifacts on the \scenename{Microphone} stand and \scenename{Material}'s object edges and ghosting artifacts in \scenename{Ship}'s mast and inside the \scenename{Lego} object. SRN produces blurry and distorted renderings in every case. Neural Volumes cannot capture the details on the \scenename{Microphone}'s grille or \scenename{Lego}'s gears, and it completely fails to recover the geometry of \scenename{Ship}'s rigging.}
\label{fig:synthresults}
\end{figure}

\paragraph{\textbf{Real images of complex scenes}}

We show results on complex real-world scenes captured with roughly forward-facing images (Table~\ref{table:results}, ``Real Forward-Facing''). This dataset consists of 8 scenes captured with a handheld cellphone (5 taken from the LLFF paper and 3 that we capture), captured with 20 to 62 images, and hold out $\nicefrac{1}{8}$ of these for the test set. All images are $1008 \times 756$ pixels.

\renewcommand{\resultsfigwidth}{1.27in}
\renewcommand{\resultscropwidth}{0.84in}

\newcommand{\cropfern}[1]{
  \makecell{
  \includegraphics[trim={774px 178px 54px 398px}, clip, width=\resultscropwidth]{#1} \\
  \includegraphics[trim={93px 339px 801px 303px}, clip, width=\resultscropwidth]{#1} 
  }
}

\newcommand{\croporchid}[1]{
  \makecell{
  \includegraphics[trim={205px 404px 681px 230px}, clip, width=\resultscropwidth]{#1} \\
  \includegraphics[trim={516px 393px 364px 235px}, clip, width=\resultscropwidth]{#1} 
  }
}

\newcommand{\cropredtrex}[1]{
  \makecell{
  \includegraphics[trim={384px 280px 478px 330px}, clip, width=\resultscropwidth]{#1}   \\
  \includegraphics[trim={183px 61px 715px 585px}, clip, width=\resultscropwidth]{#1} 
  }
}

\afterpage{\clearpage}
\begin{figure}[p]
\centering
\scriptsize
\begin{tabular}{@{}c@{\,\,}c@{}c@{}c@{}c@{}}
\makecell[c]{
\includegraphics[trim={0px 0px 0px 0px}, clip, width=\resultsfigwidth]{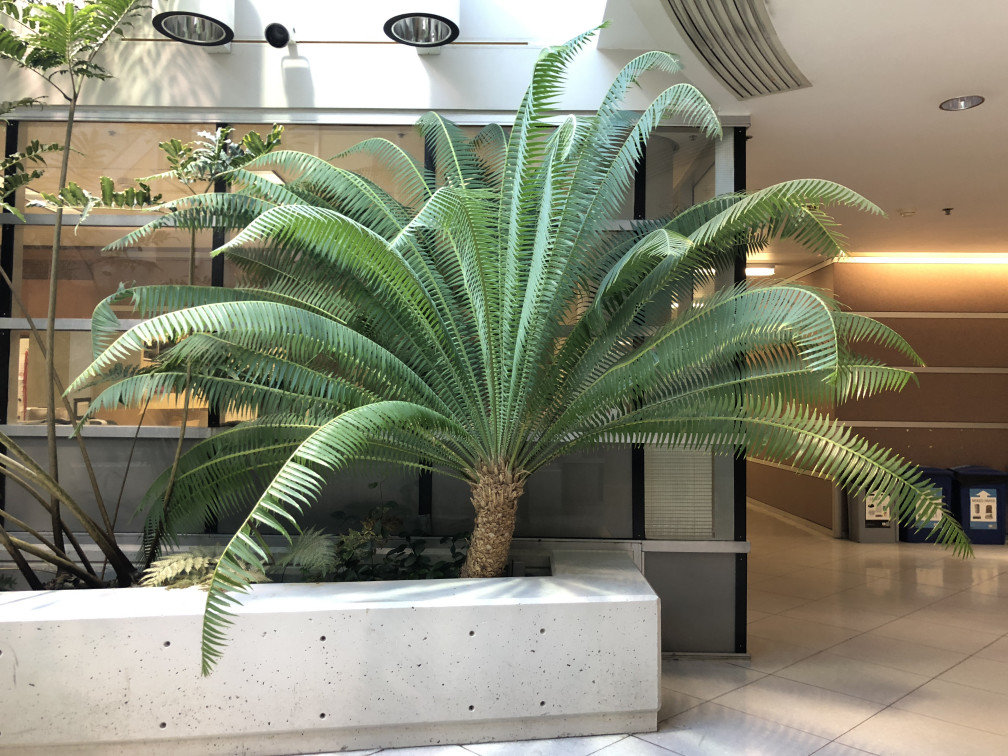}
\\
\scenename{Fern}
}
& 
\cropfern{figs/real_results_new/fern/gt.jpg} &
\cropfern{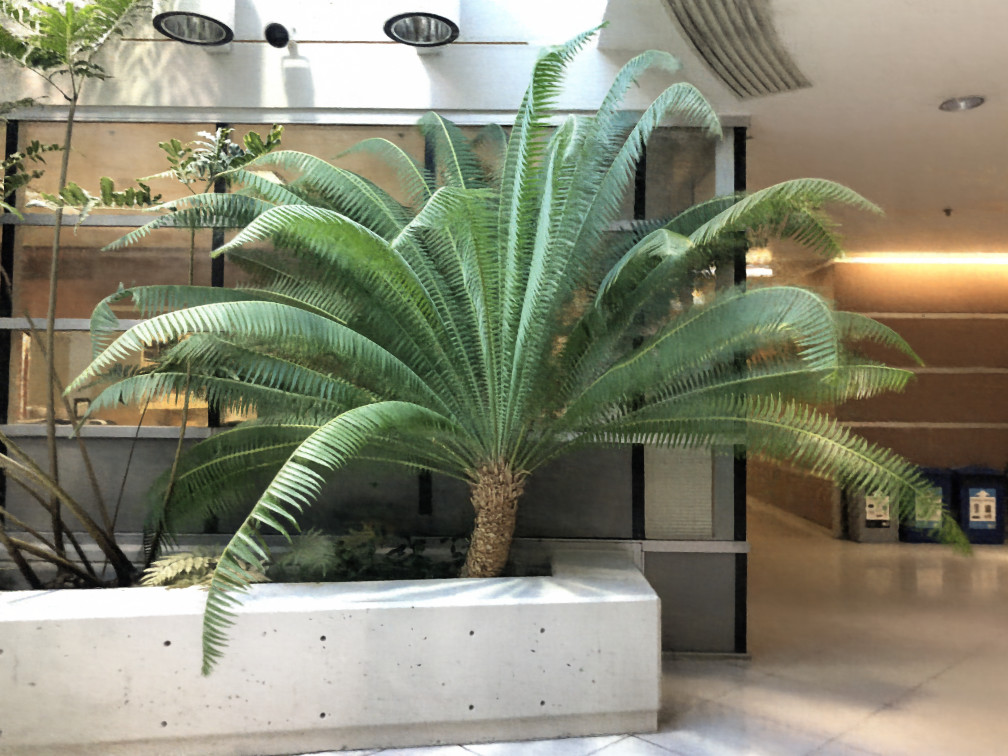} &
\cropfern{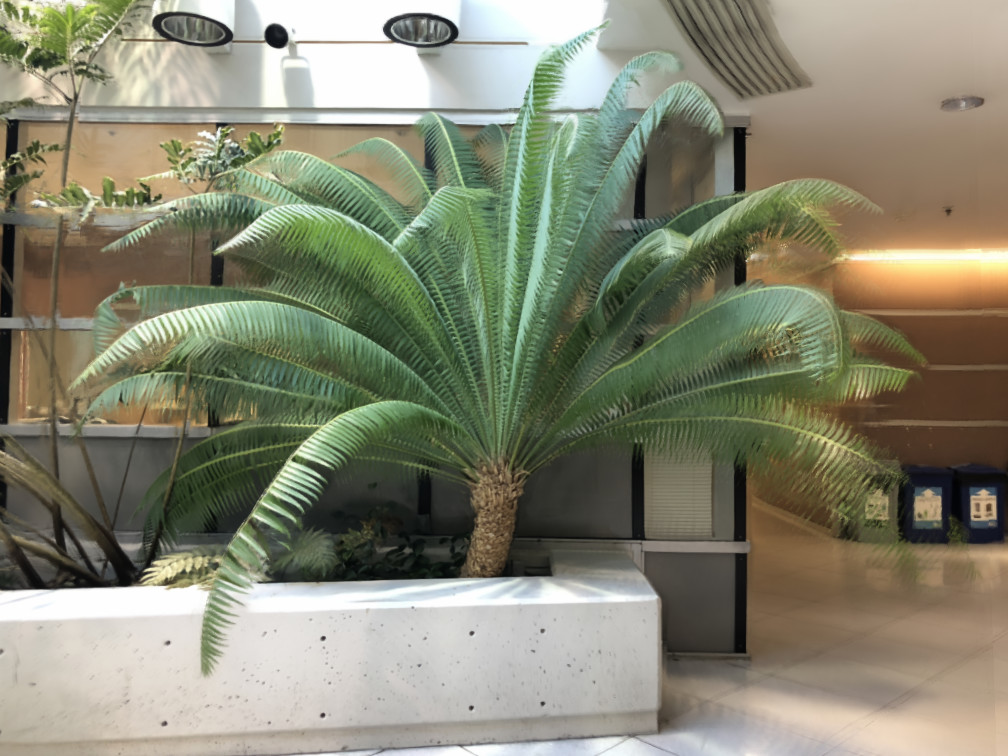} &
\cropfern{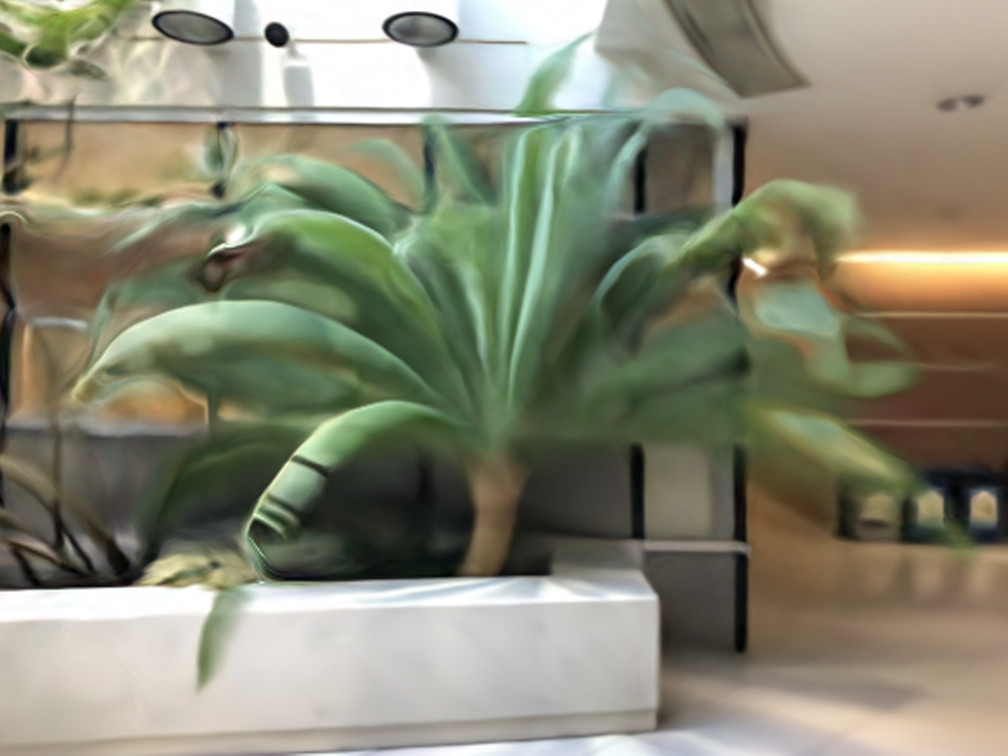} \\
\makecell[c]{
\includegraphics[trim={0px 0px 0px 0px}, clip, width=\resultsfigwidth]{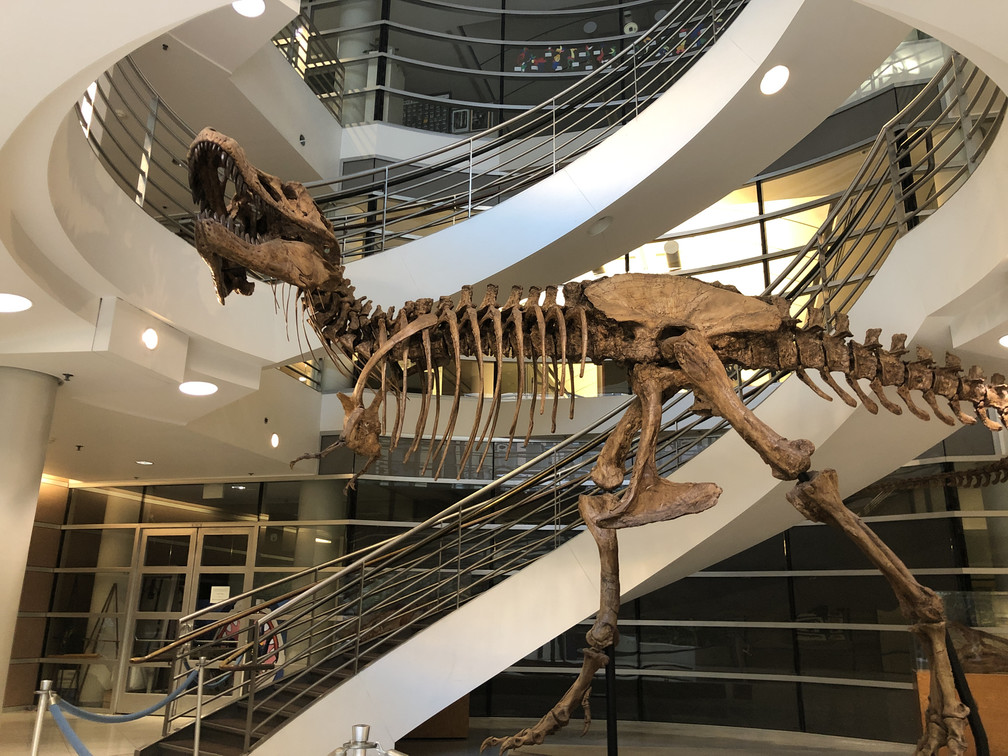}
\\
\scenename{T-Rex}
}
& 
\cropredtrex{figs/real_results_new/trex/gt.jpg} &
\cropredtrex{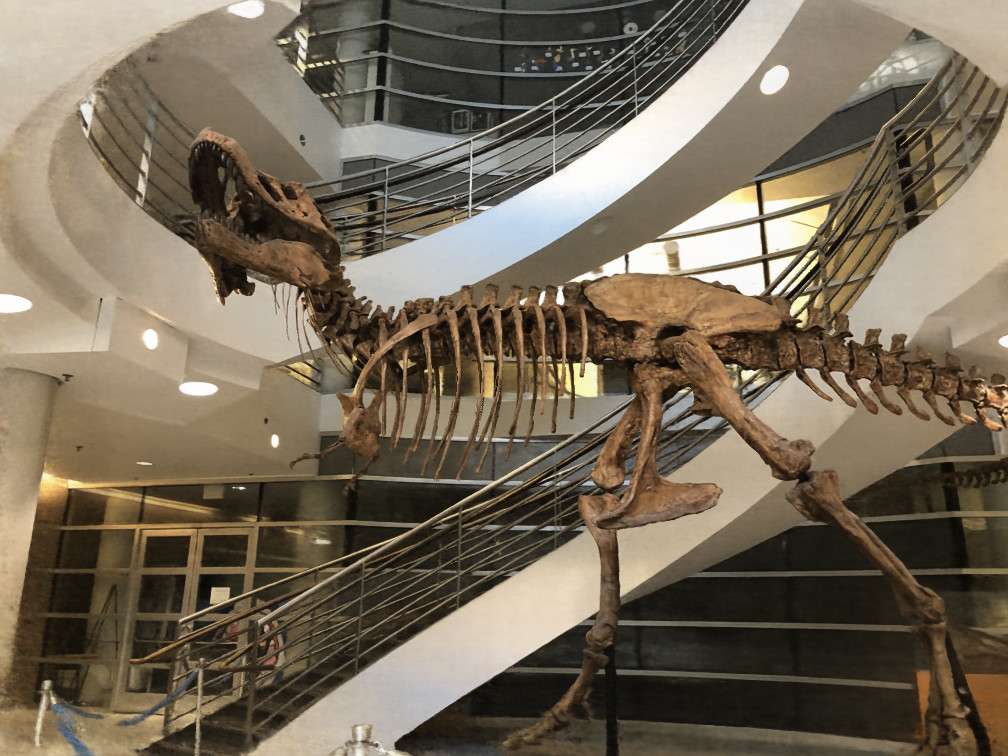} &
\cropredtrex{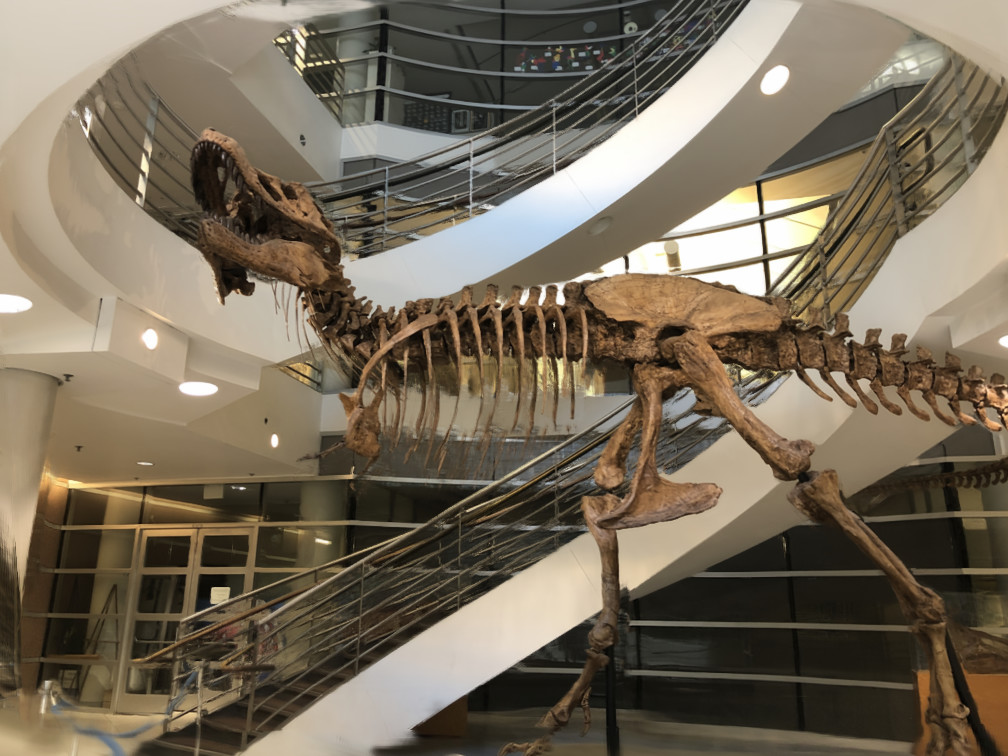} &
\cropredtrex{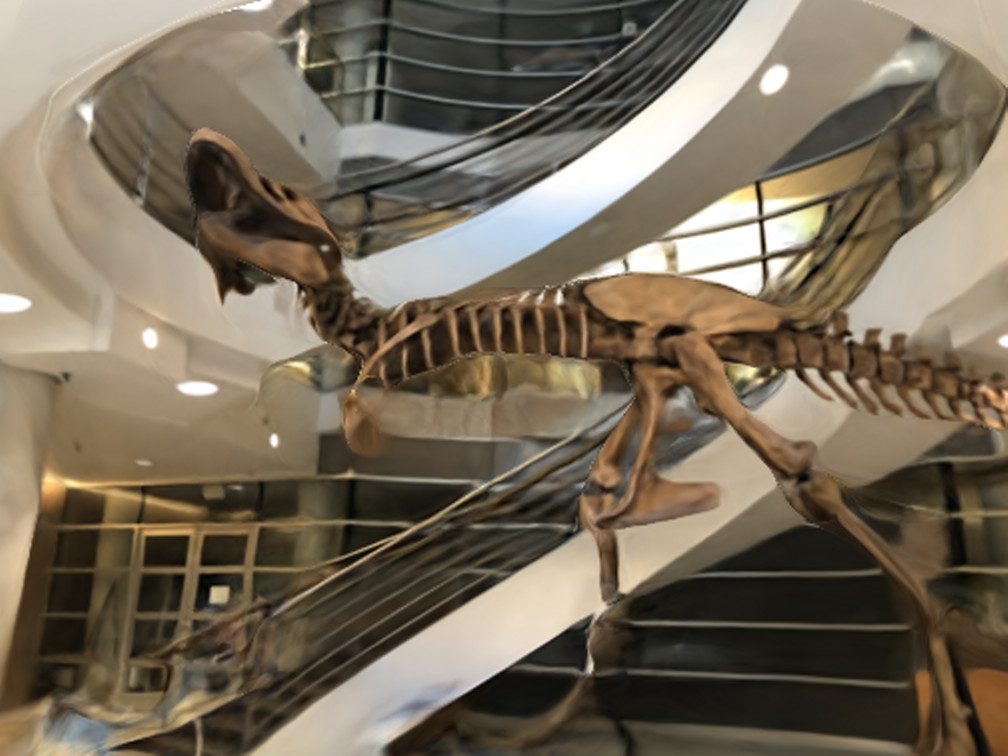} \\
\makecell[c]{
\includegraphics[trim={0px 0px 0px 0px}, clip, width=\resultsfigwidth]{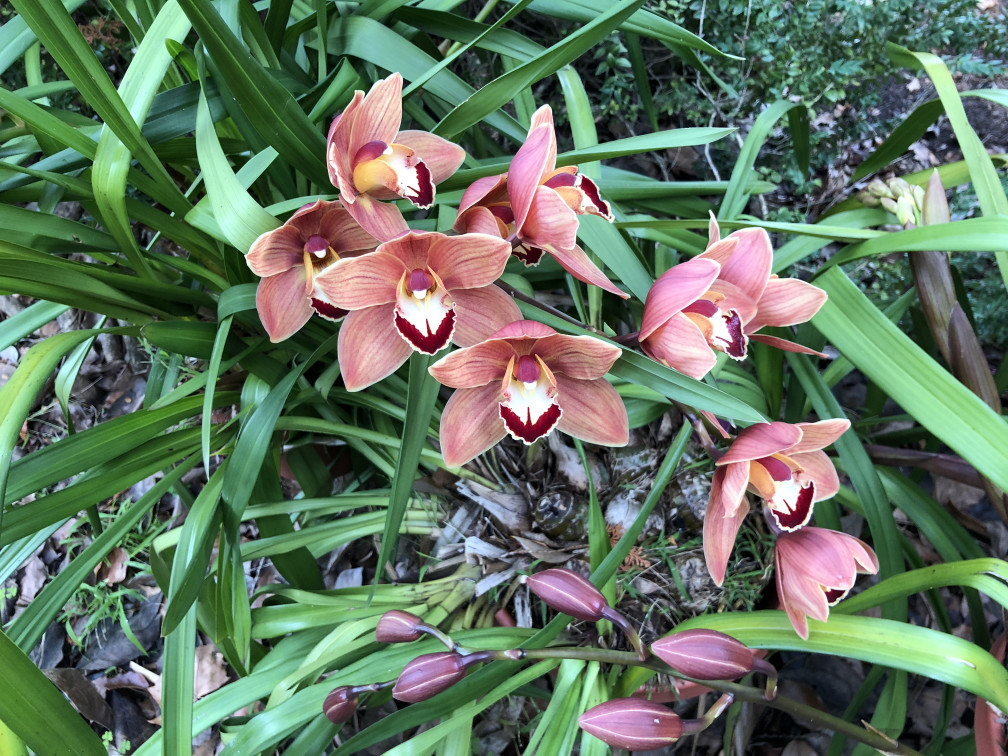}
\\
\scenename{Orchid}
}
& 
\croporchid{figs/real_results_new/orchid/gt.jpg} &
\croporchid{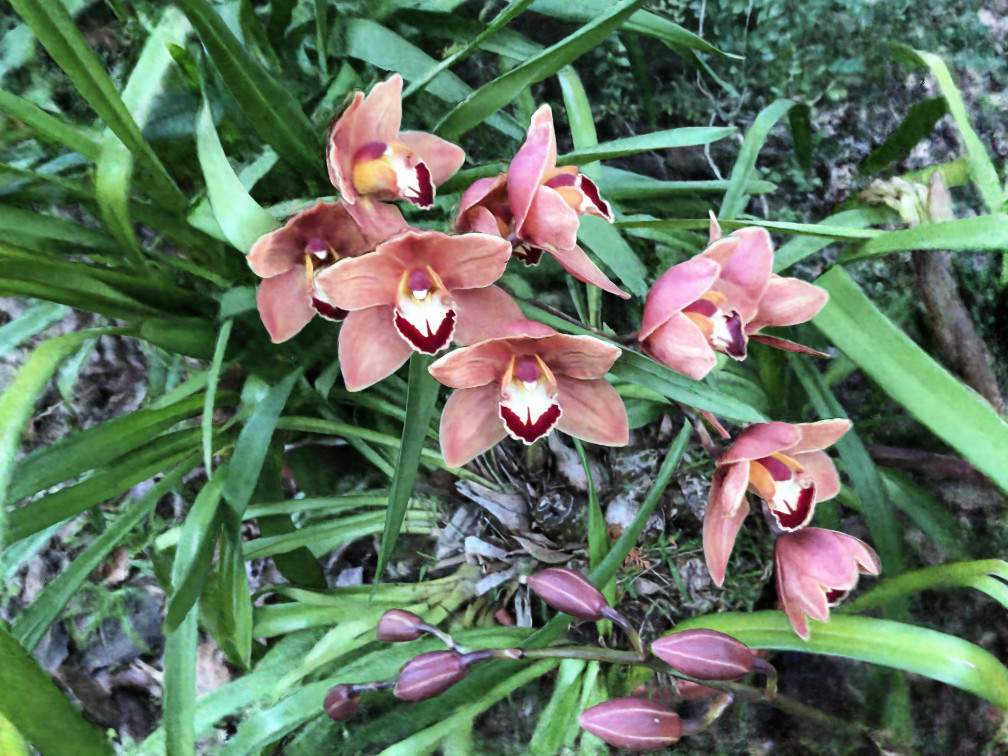} &
\croporchid{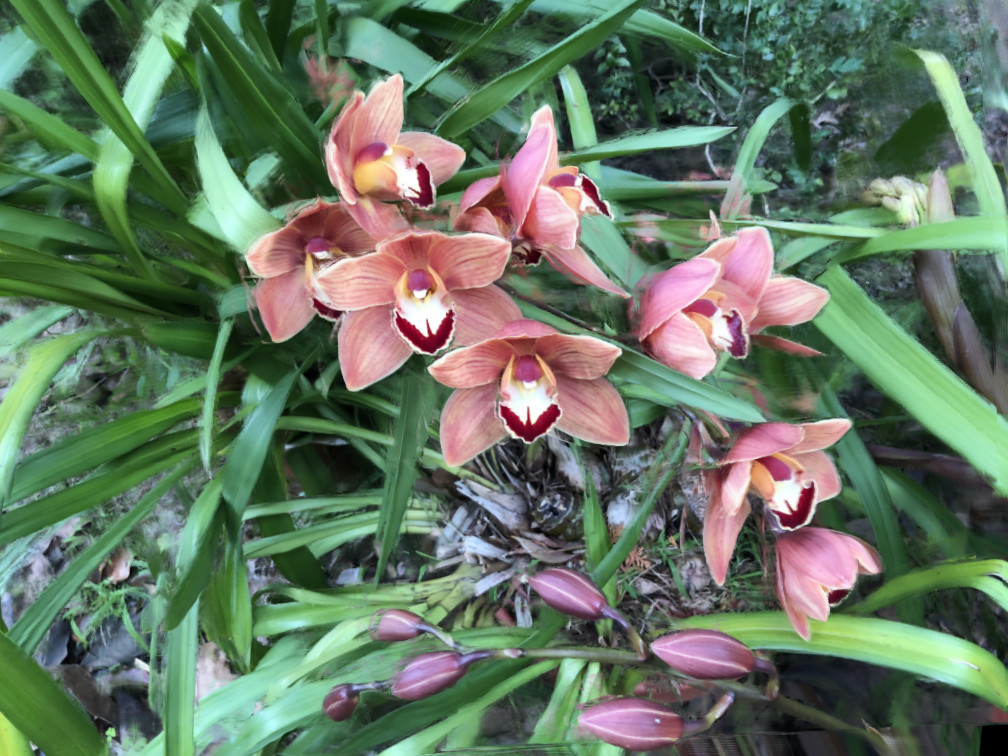} &
\croporchid{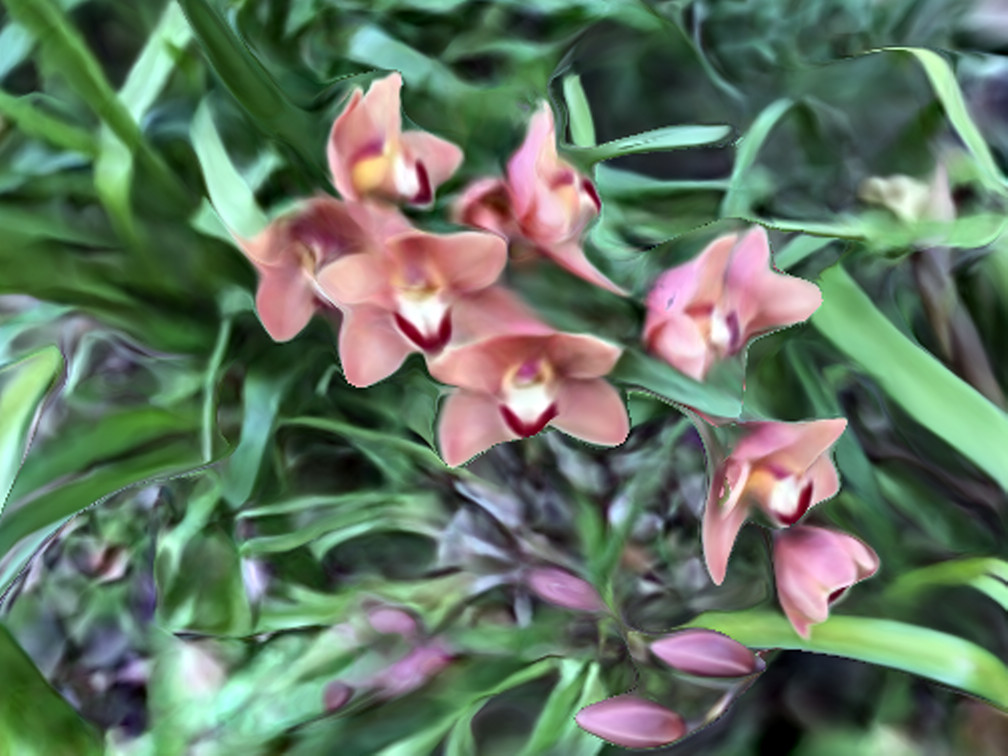} \\
& Ground Truth & NeRF (ours) & LLFF~\cite{mildenhall19} & SRN~\cite{srn} 
\end{tabular}
\caption{Comparisons on test-set views of real world scenes. LLFF is specifically designed for this use case (forward-facing captures of real scenes). Our method is able to represent fine geometry more consistently across rendered views than LLFF, as shown in \scenename{Fern}'s leaves and the skeleton ribs and railing in \scenename{T-rex}. Our method also correctly reconstructs partially occluded regions that LLFF struggles to render cleanly, such as the yellow shelves behind the leaves in the bottom \scenename{Fern} crop and green leaves in the background of the bottom \scenename{Orchid} crop. Blending between multiples renderings can also cause repeated edges in LLFF, as seen in the top \scenename{Orchid} crop. SRN captures the low-frequency geometry and color variation in each scene but is unable to reproduce any fine detail.
}
\label{fig:realresults}
\end{figure}

\newcommand{\tablespace}{\,\,\,\,}
\newcommand{\halftablespace}{\,}
\setlength{\tabcolsep}{4pt}
\begin{table}[t]
\centering
\resizebox{\textwidth}{!}{%

\begin{tabular}{l|ccc|ccc|ccc}
& \multicolumn{3}{c|}{Diffuse Synthetic $360^\circ$~\cite{deepvoxels}} & \multicolumn{3}{c|}{Realistic Synthetic $360^\circ$} & \multicolumn{3}{c}{Real Forward-Facing~\cite{mildenhall19}} \\
Method & PSNR$\uparrow$ & SSIM$\uparrow$ & LPIPS$\downarrow$ & PSNR$\uparrow$ & SSIM$\uparrow$ & LPIPS$\downarrow$ & PSNR$\uparrow$ & SSIM$\uparrow$ & LPIPS$\downarrow$ \\
\hline
SRN~\cite{srn} & $33.20$ & $0.963$ & $0.073$ & $22.26$ & $0.846$ & $0.170$ & $22.84$ & $0.668$ & $0.378$ \\
NV~\cite{neuralvolumes} & $29.62$ & $0.929$ & $0.099$ & $26.05$ & $0.893$ & $0.160$ & - & - & - \\
LLFF~\cite{mildenhall19} & $34.38$ & $0.985$ & $0.048$ & $24.88$ & $0.911$ & $0.114$ & $24.13$ & $0.798$ & $\mathbf{0.212}$\\
Ours & $\mathbf{40.15}$ & $\mathbf{0.991}$ & $\mathbf{0.023}$ & $\mathbf{31.01}$ & $\mathbf{0.947}$ & $\mathbf{0.081}$ & $\mathbf{26.50}$ & $\mathbf{0.811}$ & $0.250$\\
\end{tabular}
} \vspace{2mm}
\caption{Our method quantitatively outperforms prior work on datasets of both synthetic and real images. We report PSNR/SSIM (higher is better) and LPIPS~\cite{lpips} (lower is better). 
The DeepVoxels~\cite{deepvoxels} dataset consists of 4 diffuse objects with simple geometry. Our realistic synthetic dataset consists of pathtraced renderings of 8 geometrically complex objects with complex non-Lambertian materials. The real dataset consists of handheld forward-facing captures of 8 real-world scenes (NV cannot be evaluated on this data because it only reconstructs objects inside a bounded volume). 
Though LLFF achieves slightly better LPIPS, 
we urge readers to view our supplementary video where our method achieves better multiview consistency and produces fewer artifacts than all baselines.
} \vspace{-2mm}

\label{table:results}  
\end{table}
\setlength{\tabcolsep}{1.4pt}

\subsection{Comparisons}

To evaluate our model we compare against current top-performing techniques for view synthesis, detailed below. All methods use the same set of input views to train a separate network for each scene except Local Light Field Fusion~\cite{mildenhall19}, which trains a single 3D convolutional network on a large dataset, then uses the same trained network to process input images of new scenes at test time. 

\paragraph{\textbf{Neural Volumes (NV)}~\cite{neuralvolumes}}
synthesizes novel views of objects that lie entirely within a bounded volume in front of a distinct background (which must be separately captured without the object of interest). It optimizes a deep 3D convolutional network to predict a discretized RGB$\alpha$ voxel grid with $128^3$ samples as well as a 3D warp grid with $32^3$ samples. The algorithm renders novel views by marching camera rays through the warped voxel grid. 

\paragraph{\textbf{Scene Representation Networks (SRN)}~\cite{srn}}
represent a continuous scene as an opaque surface, implicitly defined by a MLP that maps each $(x,y,z)$ coordinate to a feature vector. They train a recurrent neural network to march along a ray through the scene representation by using the feature vector at any 3D coordinate to predict the next step size along the ray. The feature vector from the final step is decoded into a single color for that point on the surface. Note that SRN is a better-performing followup to DeepVoxels~\cite{deepvoxels} by the same authors, which is why we do not include comparisons to DeepVoxels.

\paragraph{\textbf{Local Light Field Fusion (LLFF)}~\cite{mildenhall19}}
LLFF is designed for producing photorealistic novel views for well-sampled forward facing scenes. It uses a trained 3D convolutional network to directly predict a discretized frustum-sampled RGB$\alpha$ grid (multiplane image or MPI~\cite{zhou18}) for each input view, then renders novel views by alpha compositing and blending nearby MPIs into the novel viewpoint.

\subsection{Discussion}

We thoroughly outperform both baselines that also optimize a separate network per scene (NV and SRN) in all scenarios.
Furthermore, we produce qualitatively and quantitatively superior renderings compared to LLFF (across all except one metric) while using only their input images as our entire training set.

The SRN method produces heavily smoothed geometry and texture, and its representational power for view synthesis is limited by selecting only a single depth and color per camera ray. The NV baseline is able to capture reasonably detailed volumetric geometry and appearance, but its use of an underlying explicit $128^3$ voxel grid prevents it from scaling to represent fine details at high resolutions. LLFF specifically provides a ``sampling guideline'' to not exceed 64 pixels of disparity between input views, so it frequently fails to estimate correct geometry in the synthetic datasets which contain up to 400-500 pixels of disparity between views. Additionally, LLFF blends between different scene representations for rendering different views, resulting in perceptually-distracting inconsistency as is apparent in our supplementary video. 

The biggest practical tradeoffs between these methods are time versus space. All compared single scene methods take at least 12 hours to train per scene. In contrast, LLFF can process a small input dataset in under 10 minutes. However, LLFF produces a large 3D voxel grid for every input image, resulting in enormous storage requirements (over 15GB for one ``Realistic Synthetic'' scene). Our method requires only 5 MB for the network weights (a relative compression of $3000\times$ compared to LLFF), which is even less memory than the \emph{input images alone} for a single scene from any of our datasets.

\subsection{Ablation studies}
\label{sec:ablations}

\setlength{\tabcolsep}{4pt}
\begin{table}[t]
\centering
\begin{tabular}{@{\,}l@{\,\,}l|cccc|ccc@{\,}}
& & Input & \#Im. & $\numfrequencies$ & $(\,\numsamplescoarse\,, \,\numsamplesfine\,)$ &  PSNR$\uparrow$ & SSIM$\uparrow$ & LPIPS$\downarrow$  \\
\hline
1) & No PE, VD, H & $\posxyz$ & 100 & -  & (256, \,\,-\,\,)    &  $26.67$ & $0.906$ & $0.136$  \\
2) & No Pos. Encoding & $\posall$  & 100 & -  &  (64, 128)     &  $28.77$ & $0.924$ & $0.108$  \\
3) & No View Dependence & $\posxyz$ & 100 & 10 &  (64, 128)    &  $27.66$ & $0.925$ & $0.117$  \\
4) & No Hierarchical & $\posall$ & 100 & 10 & (256, \,\,-\,\,) &  $30.06$ & $0.938$ & $0.109$  \\
\arrayrulecolor{gray}
\hline
\arrayrulecolor{black}
5) & Far Fewer Images & $\posall$  & 25  & 10 &  (64, 128)     &  $27.78$ & $0.925$ & $0.107$  \\
6) & Fewer Images  & $\posall$  & 50  & 10 &  (64, 128)        &  $29.79$ & $0.940$ & $0.096$  \\
\arrayrulecolor{gray}
\hline
\arrayrulecolor{black}
7) & Fewer Frequencies & $\posall$ & 100 & 5  &  (64, 128)     &  $30.59$ & $0.944$ & $0.088$  \\
8) & More Frequencies & $\posall$ & 100 & 15 &  (64, 128)      &  $30.81$ & $0.946$ & $0.096$  \\
\arrayrulecolor{gray}
\hline
\arrayrulecolor{black}
9) & Complete Model & $\posall$ & 100 & 10 & (64, 128)         &  $\mbf{31.01}$ & $\mbf{0.947}$ & $\mbf{0.081}$ 
\end{tabular}\vspace{2mm}
\caption{An ablation study of our model. Metrics are averaged over the 8 scenes from our realistic synthetic dataset. See Sec.~\ref{sec:ablations} for detailed descriptions.
}
\label{table:ablations}
\end{table}
\setlength{\tabcolsep}{1.4pt}

We validate our algorithm's design choices and parameters with an extensive ablation study in Table~\ref{table:ablations}. We present results on our ``Realistic Synthetic $360\degree$'' scenes. Row 9 shows our complete model as a point of reference.
Row 1 shows a minimalist version of our model without positional encoding (PE), view-dependence (VD), or hierarchical sampling (H). In rows 2--4 we remove these three components one at a time from the full model, observing that positional encoding (row 2) and view-dependence (row 3) provide the largest quantitative benefit followed by hierarchical sampling (row 4).
Rows 5--6 show how our performance decreases as the number of input images is reduced. Note that our method's performance using only 25 input images still exceeds NV, SRN, and LLFF across all metrics when they are provided with 100 images (see supplementary material).
In rows 7--8 we validate our choice of the maximum frequency $L$ used in our positional encoding for $\mathbf x$ (the maximum frequency used for $\mathbf d$ is scaled proportionally). Only using 5 frequencies reduces performance, but increasing the number of frequencies from 10 to 15 does not improve performance. We believe the benefit of increasing $L$ is limited once $2^L$ exceeds the maximum frequency present in the sampled input images (roughly 1024 in our data).

\section{Conclusion}

Our work directly addresses deficiencies of prior work that uses MLPs to represent objects and scenes as continuous functions. We demonstrate that representing scenes as 5D neural radiance fields (an MLP that outputs volume density and view-dependent emitted radiance as a function of 3D location and 2D viewing direction) produces better renderings than the previously-dominant approach of training deep convolutional networks to output discretized voxel representations. 

Although we have proposed a hierarchical sampling strategy to make rendering more sample-efficient (for both training and testing), there is still much more progress to be made in investigating techniques to efficiently optimize and render neural radiance fields. Another direction for future work is interpretability: sampled representations such as voxel grids and meshes admit reasoning about the expected quality of rendered views and failure modes, but it is unclear how to analyze these issues when we encode scenes in the weights of a deep neural network. We believe that this work makes progress towards a graphics pipeline based on real world imagery, where complex scenes could be composed of neural radiance fields optimized from images of actual objects and scenes. \\

\noindent \textbf{Acknowledgements} We thank Kevin Cao, Guowei Frank Yang, and Nithin Raghavan for comments and discussions. RR acknowledges funding from ONR grants N000141712687 and N000142012529 and the Ronald L. Graham Chair. BM is funded by a Hertz Foundation Fellowship, and MT is funded by an NSF Graduate Fellowship. Google provided a generous donation of cloud compute credits through the BAIR Commons program. We thank the following Blend Swap users for the models used in our realistic synthetic dataset: gregzaal (ship), 1DInc (chair), bryanajones (drums), Herberhold (ficus), erickfree (hotdog), Heinzelnisse (lego), elbrujodelatribu (materials), and up3d.de (mic).

\bibliographystyle{splncs04}
\bibliography{arxiv_submission}

\appendix

\section{Additional Implementation Details}

\begin{figure}[t]
\centering
\includegraphics[width=0.8\linewidth]{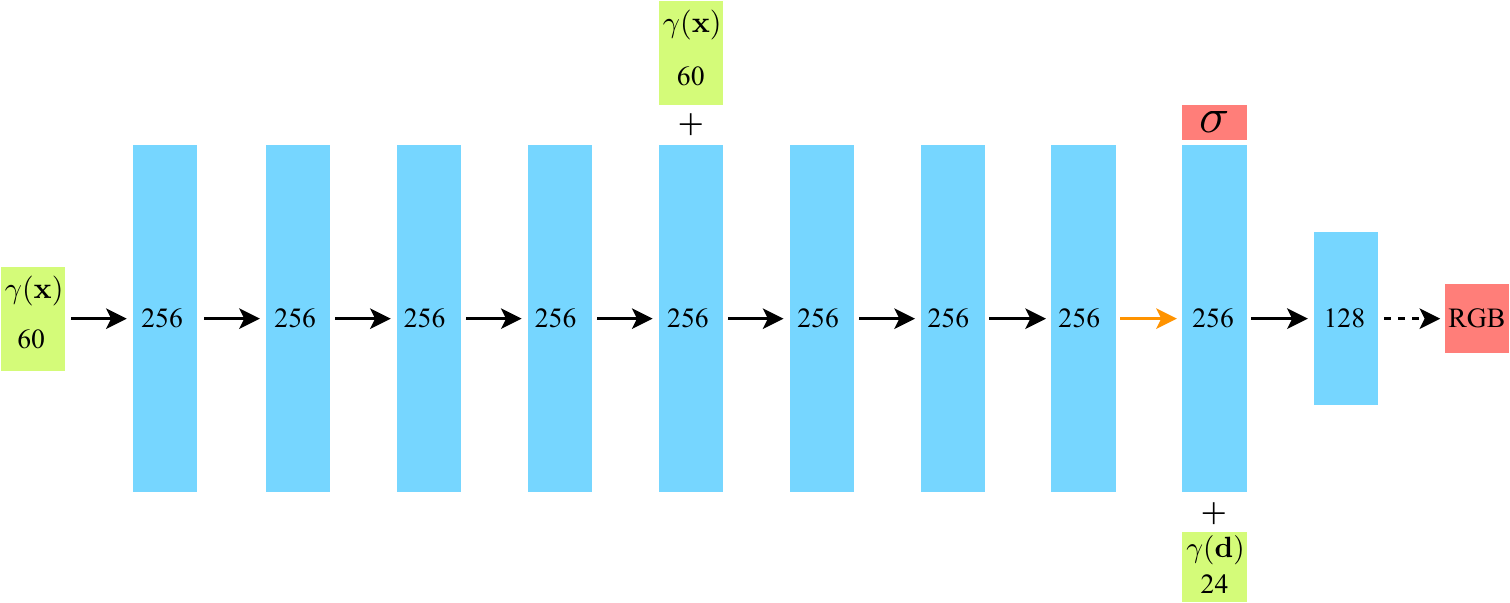}
\caption{A visualization of our fully-connected network architecture. Input vectors are shown in green, intermediate hidden layers are shown in blue, output vectors are shown in red, and the number inside each block signifies the vector's dimension. All layers are standard fully-connected layers, black arrows indicate layers with ReLU activations, orange arrows indicate layers with no activation, dashed black arrows indicate layers with sigmoid activation, and ``+'' denotes vector concatenation. The positional encoding of the input location ($\gamma(\mathbf{x})$) is passed through 8 fully-connected ReLU layers, each with 256 channels. We follow the DeepSDF~\cite{deepsdf} architecture and include a skip connection that concatenates this input to the fifth layer's activation. An additional layer outputs the volume density $\sigma$ (which is rectified using a ReLU to ensure that the output volume density is nonnegative) and a 256-dimensional feature vector. This feature vector is concatenated with the positional encoding of the input viewing direction ($\gamma(\mathbf{d})$), and is processed by an additional fully-connected ReLU layer with 128 channels. A final layer (with a sigmoid activation) outputs the emitted RGB radiance at position $\mathbf{x}$, as viewed by a ray with direction $\mathbf{d}$.}
\label{fig:net}
\end{figure}

\paragraph{\textbf{Network Architecture}}  
Fig.~\ref{fig:net} details our simple fully-connected architecture.

\paragraph{\textbf{Volume Bounds}}
Our method renders views by querying the neural radiance field representation at continuous 5D coordinates along camera rays. For experiments with synthetic images, we scale the scene so that it lies within a cube of side length 2 centered at the origin, and only query the representation within this bounding volume. Our dataset of real images contains content that can exist anywhere between the closest point and infinity, so we use normalized device coordinates to map the depth range of these points into $[-1,1]$. This shifts all the ray origins to the near plane of the scene, maps the perspective rays of the camera to parallel rays in the transformed volume, and uses disparity (inverse depth) instead of metric depth, so all coordinates are now bounded.

\paragraph{\textbf{Training Details}}
For real scene data, we regularize our network by adding random Gaussian noise with zero mean and unit variance to the output $\sigma$ values (before passing them through the ReLU) during optimization, finding that this slightly improves visual performance for rendering novel views. We implement our model in Tensorflow~\cite{tensorflow}.

\paragraph{\textbf{Rendering Details}}
To render new views at test time, we sample $64$ points per ray through the coarse network and $64+128=192$ points per ray through the fine network, for a total of $256$ network queries per ray. Our realistic synthetic dataset requires 640k rays per image, and our real scenes require 762k rays per image, resulting in between 150 and 200 million network queries per rendered image. On an NVIDIA V100, this takes approximately 30 seconds per frame.

\section{Additional Baseline Method Details}

\paragraph{\textbf{Neural Volumes (NV)}~\cite{neuralvolumes}}
We use the NV code open-sourced by the authors at \url{https://github.com/facebookresearch/neuralvolumes} and follow their procedure for training on a single scene without time dependence. 

\paragraph{\textbf{Scene Representation Networks (SRN)}~\cite{srn}}
We use the SRN code open-sourced by the authors at \url{https://github.com/vsitzmann/scene-representation-networks} and follow their procedure for training on a single scene.

\paragraph{\textbf{Local Light Field Fusion (LLFF)}~\cite{mildenhall19}}
We use the pretrained LLFF model open-sourced by the authors at \url{https://github.com/Fyusion/LLFF}.

\paragraph{\textbf{Quantitative Comparisons}} The SRN implementation published by the authors requires a significant amount of GPU memory, and is limited to an image resolution of $512 \times 512$ pixels even when parallelized across 4 NVIDIA V100 GPUs. We compute quantitative metrics for SRN at $512 \times 512$ pixels for our synthetic datasets and $504 \times 376$ pixels for the real datasets, in comparison to $800\times 800$ and $1008 \times 752$ respectively for the other methods that can be run at higher resolutions.

\section{NDC ray space derivation}

We reconstruct real scenes with ``forward facing'' captures in the normalized device coordinate (NDC) space that is commonly used as part of the triangle rasterization pipeline. This space is convenient because it preserves parallel lines while converting the $z$ axis (camera axis) to be linear in disparity.

Here we derive the transformation which is applied to rays to map them from camera space to NDC space. The standard 3D perspective projection matrix for homogeneous coordinates is:
\begin{equation}
    M = \begin{pmatrix}
     \frac{n}{r} & 0 & 0 & 0 \\
     0 & \frac{n}{t} & 0 & 0 \\
     0 & 0 & \frac{-(f+n)}{f-n} & \frac{-2fn}{f-n} \\
     0 & 0 & -1 & 0 
    \end{pmatrix}
\end{equation}
where $n, f$ are the near and far clipping planes and $r$ and $t$ are the right and top bounds of the scene at the near clipping plane. (Note that this is in the convention where the camera is looking in the $-z$ direction.) To project a homogeneous point $(x,y,z,1)^\top$, we left-multiply by M and then divide by the fourth coordinate:
\begin{align}
    \begin{pmatrix}
     \frac{n}{r} & 0 & 0 & 0 \\
     0 & \frac{n}{t} & 0 & 0 \\
     0 & 0 & \frac{-(f+n)}{f-n} & \frac{-2fn}{f-n} \\
     0 & 0 & -1 & 0 
    \end{pmatrix}
    \begin{pmatrix}
    x \\ y \\ z \\ 1
    \end{pmatrix}
    &=
    \begin{pmatrix}
    \frac{n}{r} x \\
    \frac{n}{t} y \\
    \frac{-(f+n)}{f-n} z - \frac{-2fn}{f-n} \\
    -z
    \end{pmatrix} \\
    \textrm{project}
    &\rightarrow
    \begin{pmatrix}
    \frac{n}{r} \frac{x}{-z} \\
    \frac{n}{t} \frac{y}{-z} \\
    \frac{(f+n)}{f-n} - \frac{2fn}{f-n} \frac{1}{-z}
    \end{pmatrix}
    \label{eq:projpt}
\end{align}
The projected point is now in normalized device coordinate (NDC) space, where the original viewing frustum has been mapped to the cube $[-1,1]^3$. 

Our goal is to take a ray $\mathbf o + t \mathbf d$ and calculate a ray origin $\mathbf o'$ and direction $\mathbf d'$ in NDC space such that for every $t$, there exists a new $t'$ for which $\pi(\mathbf o + t \mathbf d) = \mathbf o' + t' \mathbf d'$ (where $\pi$ is projection using the above matrix). In other words, the projection of the original ray and the NDC space ray trace out the same points (but not necessarily at the same rate). 

Let us rewrite the projected point from Eqn.~\ref{eq:projpt} as $(a_x x/z, a_y y/z, a_z + b_z / z)^\top$. The components of the new origin $\mathbf o'$ and direction $\mathbf d'$ must satisfy:
\begin{align}
    \begin{pmatrix}
        a_x \frac{o_x + t d_x}{o_z + t d_z} \\[6pt]
        a_y \frac{o_y + t d_y}{o_z + t d_z} \\[6pt]
        a_z + \frac{b_z}{o_z + t d_z}
    \end{pmatrix}
    =
    \begin{pmatrix}
        o_x' + t' d_x' \\
        o_y' + t' d_y' \\
        o_z' + t' d_z' 
    \end{pmatrix} \, .
    \label{eq:rayeq}
\end{align}
To eliminate a degree of freedom, we decide that $t'=0$ and $t=0$ should map to the same point. Substituting $t=0$ and $t'=0$ Eqn.~\ref{eq:rayeq} directly gives our NDC space origin $\mathbf o'$:
\begin{align}
    \mathbf o' =
    \begin{pmatrix}
        o_x' \\
        o_y' \\
        o_z'
    \end{pmatrix}
    =
    \begin{pmatrix}
        a_x \frac{o_x}{o_z} \\[6pt]
        a_y \frac{o_y}{o_z} \\[6pt]
        a_z + \frac{b_z}{o_z}
    \end{pmatrix}
    = \pi(\mathbf o) \,. 
\end{align}
This is exactly the projection $\pi(\mathbf o)$ of the original ray's origin. By substituting this back into Eqn.~\ref{eq:rayeq} for arbitrary $t$, we can determine the values of $t'$ and $\mathbf d'$:
\begin{align}
    \begin{pmatrix}
        t' d_x' \\
        t' d_y' \\
        t' d_z' 
    \end{pmatrix}
    &=
    \begin{pmatrix}
        a_x \frac{o_x + t d_x}{o_z + t d_z} - a_x \frac{o_x}{o_z} \\[6pt]
        a_y \frac{o_y + t d_y}{o_z + t d_z} - a_y \frac{o_y}{o_z} \\[6pt]
        a_z + \frac{b_z}{o_z + t d_z} - a_z - \frac{b_z}{o_z}
    \end{pmatrix} \\[6pt]
    &=
    \begin{pmatrix}
        a_x \frac{o_z(o_x + t d_x) - o_x(o_z + t d_z)}{(o_z + t d_z)o_z} \\[6pt]
        a_y \frac{o_z(o_y + t d_y) - o_y(o_z + t d_z)}{(o_z + t d_z)o_z} \\[6pt]
        b_z\frac{o_z - (o_z + t d_z)}{(o_z + t d_z)o_z}
    \end{pmatrix} \\[6pt]
    &= 
    \begin{pmatrix}
        a_x \frac{t d_z}{o_z + t d_z} \left(\frac{d_x}{d_z} - \frac{o_x}{o_z}\right) \\[6pt]
        a_y \frac{t d_z}{o_z + t d_z} \left(\frac{d_y}{d_z} - \frac{o_y}{o_z}\right) \\[6pt]
        -b_z \frac{t d_z}{o_z + t d_z} \frac{1}{o_z}
    \end{pmatrix} 
\end{align}
Factoring out a common expression that depends only on $t$ gives us:
\begin{align}
    t' &= \frac{t d_z}{o_z + t d_z} = 1 - \frac{o_z}{o_z + t d_z} \\[6pt]
    \mathbf d' &= 
    \begin{pmatrix}
        a_x \left(\frac{d_x}{d_z} - \frac{o_x}{o_z}\right) \\[6pt]
        a_y \left(\frac{d_y}{d_z} - \frac{o_y}{o_z}\right) \\[6pt]
        -b_z \frac{1}{o_z}
    \end{pmatrix} \,. 
\end{align}
Note that, as desired, $t'=0$ when $t=0$. Additionally, we see that $t' \to 1$ as $t \to \infty$. Going back to the original projection matrix, our constants are:
\begin{align}
    a_x &= -\frac{n}{r} \\
    a_y &= -\frac{n}{t} \\
    a_z &= \frac{f+n}{f-n} \\
    b_z &= \frac{2fn}{f-n} 
\end{align}
Using the standard pinhole camera model, we can reparameterize as:
\begin{align}
    a_x &= -\frac{f_{cam}}{W/2} \\
    a_y &= -\frac{f_{cam}}{H/2} 
\end{align}
where $W$ and $H$ are the width and height of the image in pixels and $f_{cam}$ is the focal length of the camera. 

In our real forward facing captures, we assume that the far scene bound is infinity (this costs us very little since NDC uses the $z$ dimension to represent \emph{inverse} depth, i.e., disparity). In this limit the $z$ constants simplify to:
\begin{align}
    a_z &= 1 \\
    b_z &= 2n \, .
\end{align}
Combining everything together:
\begin{align}
    \mathbf o' &= 
    \begin{pmatrix}
        -\frac{f_{cam}}{W/2} \frac{o_x}{o_z} \\[6pt]
        -\frac{f_{cam}}{H/2} \frac{o_y}{o_z} \\[6pt]
        1 + \frac{2n}{o_z}
    \end{pmatrix}
    \\[6pt]
    \mathbf d' &= 
    \begin{pmatrix}
        -\frac{f_{cam}}{W/2} \left(\frac{d_x}{d_z} - \frac{o_x}{o_z}\right) \\[6pt]
        -\frac{f_{cam}}{H/2} \left(\frac{d_y}{d_z} - \frac{o_y}{o_z}\right) \\[6pt]
        -2n \frac{1}{o_z}
    \end{pmatrix} \, .
\end{align}
One final detail in our implementation: we shift $\mathbf o$ to the ray's intersection with the near plane at $z=-n$ (before this NDC conversion) by taking $\mathbf o_n = \mathbf o + t_n \mathbf d$ for $t_n = -(n + o_z) / d_z$. Once we convert to the NDC ray, this allows us to simply sample $t'$ linearly from $0$ to $1$ in order to get a linear sampling in disparity from $n$ to $\infty$ in the original space.

\section{Additional Results}

\paragraph{\textbf{Per-scene breakdown}}
Tables~\ref{table:suppresults1},~\ref{table:suppresults2},~\ref{table:suppresults3}, and~\ref{table:suppresultsabl} include a breakdown of the quantitative results presented in the main paper into per-scene metrics. The per-scene breakdown is consistent with the aggregate quantitative metrics presented in the paper, where our method quantitatively outperforms all baselines. Although LLFF achieves slightly better LPIPS metrics, we urge readers to view our supplementary video where our method achieves better multiview consistency and produces fewer artifacts than all baselines.

\newcommand{\suppresultsfigwidth}{1.14in}
\newcommand{\suppresultscropwidth}{0.69in}

\newcommand{\cropgreek}[1]{
  \makecell{
  \includegraphics[trim={202px 201px 202px 203px}, clip, width=\suppresultscropwidth]{#1} \\
  \includegraphics[trim={221px 132px 199px 288px}, clip, width=\suppresultscropwidth]{#1}
  }
}

\newcommand{\cropcube}[1]{
  \makecell{
  \includegraphics[trim={300px 309px 136px 127px}, clip, width=\suppresultscropwidth]{#1} \\
  \includegraphics[trim={120px 132px 316px 304px}, clip, width=\suppresultscropwidth]{#1} 
  }
}

\begin{figure}[t]
\centering
\scriptsize
\begin{tabular}{@{}c@{}c@{}c@{}c@{}c@{}c@{}}
\makecell[c]{
\includegraphics[trim={0px 0px 0px 100px}, clip, width=\suppresultsfigwidth]{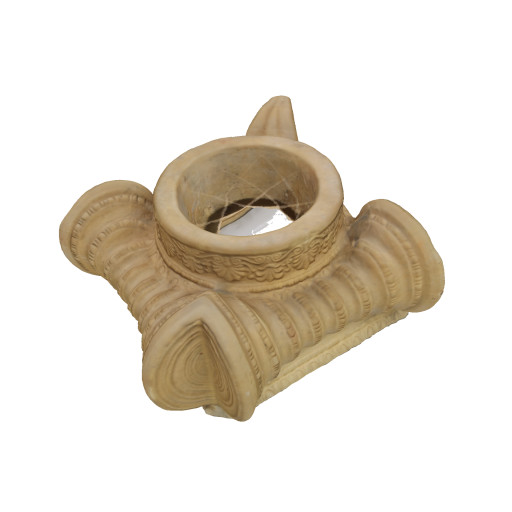}
\\
\scenename{Pedestal}
}
&
\cropgreek{figs/dvox_images/gt_greek_575.jpg} &
\cropgreek{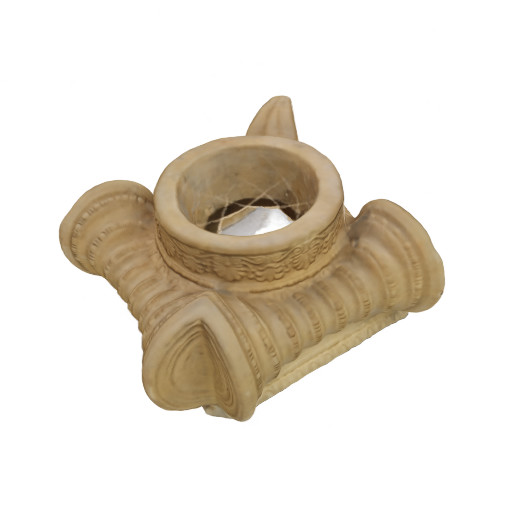} &
\cropgreek{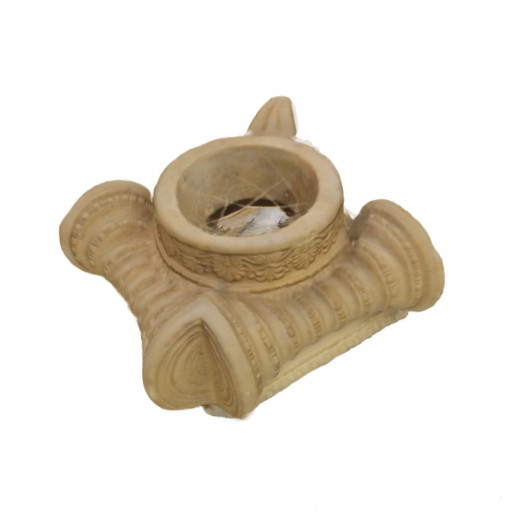} &
\cropgreek{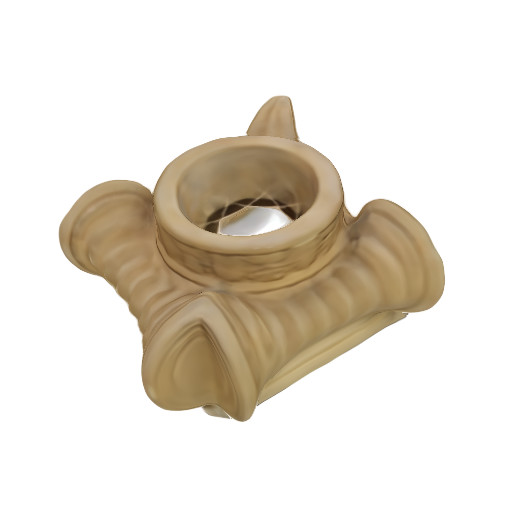} &
\cropgreek{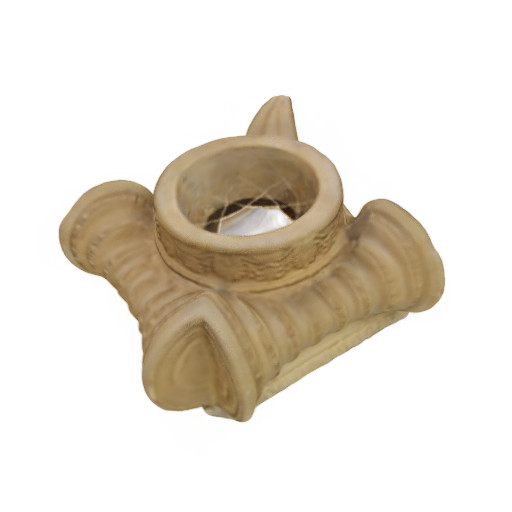} \\
\makecell[c]{
\includegraphics[trim={0px 0px 0px 100px}, clip, width=\suppresultsfigwidth]{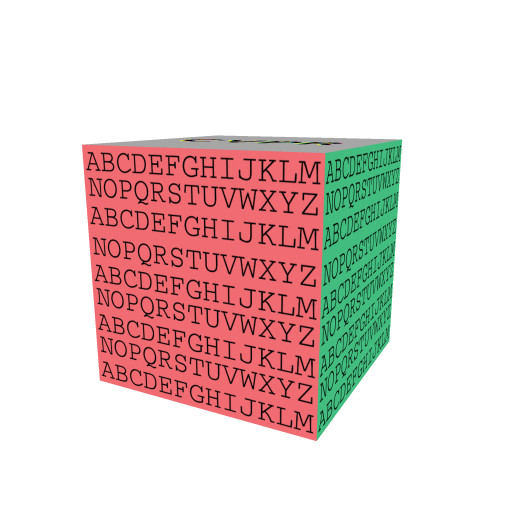}
\\
\scenename{Cube}
}
&
\cropcube{figs/dvox_images/gt_cube_160.jpg} &
\cropcube{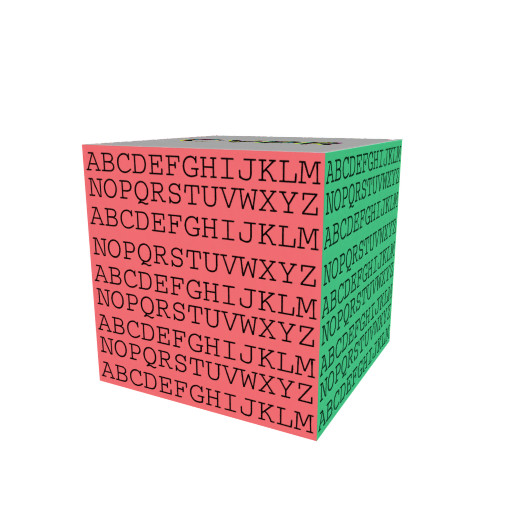} &
\cropcube{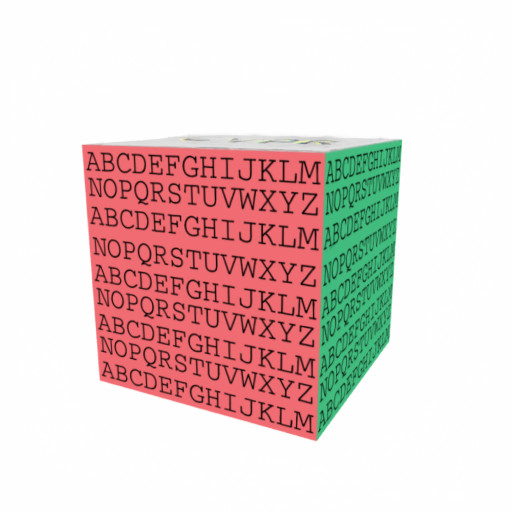} &
\cropcube{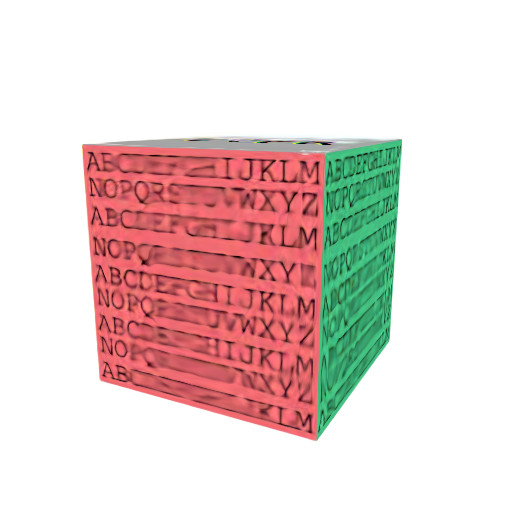} &
\cropcube{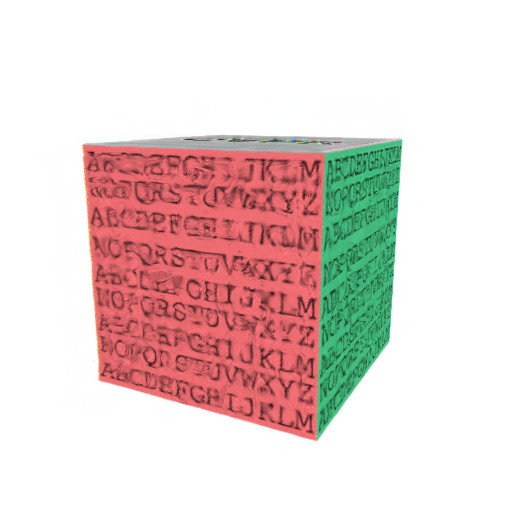} \\
& Ground Truth & NeRF (ours) & LLFF~\cite{mildenhall19} & SRN~\cite{srn} & NV~\cite{neuralvolumes}
\end{tabular} 
\caption{Comparisons on test-set views for scenes from the DeepVoxels~\cite{deepvoxels} synthetic dataset. The objects in this dataset have simple geometry and perfectly diffuse reflectance. Because of the large number of input images (479 views) and simplicity of the rendered objects, both our method and LLFF~\cite{mildenhall19} perform nearly perfectly on this data. LLFF still occasionally presents artifacts when interpolating between its 3D volumes, as in the top inset for each object. SRN~\cite{srn} and NV~\cite{neuralvolumes} do not have the representational power to render fine details.}
\label{fig:synthresults}
\end{figure}

\renewcommand{\tablespace}{\,\,\,\,}
\renewcommand{\halftablespace}{\,}

\setlength{\tabcolsep}{4pt}
\begin{table}[t]
\centering
\resizebox{\textwidth}{!}{%
\begin{tabular}{l|cccc|cccc|cccc}
& \multicolumn{4}{c|}{PSNR$\uparrow$} &
\multicolumn{4}{c|}{SSIM$\uparrow$} & \multicolumn{4}{c}{LPIPS$\downarrow$} \\
 & Chair & Pedestal & Cube & Vase & Chair & Pedestal & Cube & Vase & Chair & Pedestal & Cube & Vase \\
\hline
DeepVoxels~\cite{deepvoxels} & $33.45$ & $32.35$ & $28.42$ & $27.99$ & $0.99$ & $0.97$ & $0.97$ & $0.96$ & $-$ & $-$ & $-$ & $-$ \\
SRN~\cite{srn}           & $36.67$ & $35.91$ & $28.74$ & $31.46$ & $0.982$ & $0.957$ & $0.944$ & $0.969$ & $0.093$ & $0.081$ & $0.074$ & $0.044$ \\
NV~\cite{neuralvolumes}  & $35.15$ & $36.47$ & $26.48$ & $20.39$ & $0.980$ & $0.963$ & $0.916$ & $0.857$ & $0.096$ & $0.069$ & $0.113$ & $0.117$ \\
LLFF~\cite{mildenhall19} & $36.11$ & $35.87$ & $32.58$ & $32.97$ & $\mathbf{0.992}$ & $0.983$ & $0.983$ & $0.983$ & $0.051$ & $0.039$ & $0.064$ & $0.039$ \\
Ours       & $\mathbf{42.65}$ & $\mathbf{41.44}$ & $\mathbf{39.19}$ & $\mathbf{37.32}$ & ${0.991}$ & $\mathbf{0.986}$ & $\mathbf{0.996}$ & $\mathbf{0.992}$ & $\mathbf{0.047}$ & $\mathbf{0.024}$ & $\mathbf{0.006}$ & $\mathbf{0.017}$ \\
\end{tabular}
} \vspace{2mm}
\caption{Per-scene quantitative results from the DeepVoxels~\cite{deepvoxels} dataset. The ``scenes'' in this dataset are all diffuse objects with simple geometry, rendered from texture-mapped meshes captured by a 3D scanner. The metrics for the DeepVoxels method are taken directly from their paper, which does not report LPIPS and only reports two significant figures for SSIM.
}
\label{table:suppresults1}  
\end{table}
\setlength{\tabcolsep}{1.4pt}

\setlength{\tabcolsep}{4pt}
\begin{table}[t]
\centering
\resizebox{\textwidth}{!}{
\begin{tabular}{l|cccccccc}
\multicolumn{9}{c}{PSNR$\uparrow$}\\
& Chair & Drums & Ficus & Hotdog & Lego & Materials & Mic & Ship \\
\hline
SRN~\cite{srn}            & $26.96$ & $17.18$ & $20.73$ & $26.81$ & $20.85$ & $18.09$ & $26.85$ & $20.60$ \\
NV~\cite{neuralvolumes}   & $28.33$ & $22.58$ & $24.79$ & $30.71$ & $26.08$ & $24.22$ & $27.78$ & $23.93$ \\
LLFF~\cite{mildenhall19}  & $28.72$ & $21.13$ & $21.79$ & $31.41$ & $24.54$ & $20.72$ & $27.48$ & $23.22$ \\
Ours                      & $\mathbf{33.00}$ & $\mathbf{25.01}$ & $\mathbf{30.13}$ & $\mathbf{36.18}$ & $\mathbf{32.54}$ & $\mathbf{29.62}$ & $\mathbf{32.91}$ & $\mathbf{28.65}$ \\
\hline
\multicolumn{9}{c}{}
\end{tabular}
} \vspace{2mm}
\resizebox{\textwidth}{!}{
\begin{tabular}{l|cccccccc}
\multicolumn{9}{c}{SSIM$\uparrow$}\\
& Chair & Drums & Ficus & Hotdog & Lego & Materials & Mic & Ship \\
\hline
SRN~\cite{srn}            & $0.910$ & $0.766$ & $0.849$ & $0.923$ & $0.809$ & $0.808$ & $0.947$ & $0.757$ \\
NV~\cite{neuralvolumes}   & $0.916$ & $0.873$ & $0.910$ & $0.944$ & $0.880$ & $0.888$ & $0.946$ & $0.784$ \\
LLFF~\cite{mildenhall19}  & $0.948$ & $0.890$ & $0.896$ & $0.965$ & $0.911$ & $0.890$ & $0.964$ & $0.823$ \\
Ours                      & $\mbf{0.967}$ & $\mbf{0.925}$ & $\mbf{0.964}$ & $\mbf{0.974}$ & $\mbf{0.961}$ & $\mbf{0.949}$ & $\mbf{0.980}$ & $\mbf{0.856}$ \\
\hline
\multicolumn{9}{c}{}
\end{tabular}
} \vspace{2mm}
\resizebox{\textwidth}{!}{
\begin{tabular}{l|cccccccc}
\multicolumn{9}{c}{LPIPS$\downarrow$}\\
& Chair & Drums & Ficus & Hotdog & Lego & Materials & Mic & Ship \\
\hline
SRN~\cite{srn}            & $0.106$ & $0.267$ & $0.149$ & $0.100$ & $0.200$ & $0.174$ & $0.063$ & $0.299$ \\
NV~\cite{neuralvolumes}   & $0.109$ & $0.214$ & $0.162$ & $0.109$ & $0.175$ & $0.130$ & $0.107$ & $0.276$ \\
LLFF~\cite{mildenhall19}  & $0.064$ & $0.126$ & $0.130$ & $\mathbf{0.061}$ & $0.110$ & $0.117$ & $0.084$ & $0.218$ \\
Ours                      & $\mathbf{0.046}$ & $\mathbf{0.091}$ & $\mathbf{0.044}$ & ${0.121}$ & $\mathbf{0.050}$ & $\mathbf{0.063}$ & $\mathbf{0.028}$ & $\mathbf{0.206}$ \\
\hline
\end{tabular}
} \vspace{2mm}
\caption{Per-scene quantitative results from our realistic synthetic dataset. The ``scenes'' in this dataset are all objects with more complex gometry and non-Lambertian materials, rendered using Blender's Cycles pathtracer.
}
\label{table:suppresults2} 
\end{table}
\setlength{\tabcolsep}{1.4pt}


\setlength{\tabcolsep}{4pt}
\begin{table}[t]
\centering
\resizebox{\textwidth}{!}{
\begin{tabular}{l|cccccccc}
\multicolumn{9}{c}{PSNR$\uparrow$}\\
& Room & Fern & Leaves & Fortress & Orchids & Flower & T-Rex & Horns \\
\hline
SRN~\cite{srn}            & $27.29$ & $21.37$ & $18.24$ & $26.63$ & $17.37$ & $24.63$ & $22.87$ & $24.33$ \\
LLFF~\cite{mildenhall19}  & $28.42$ & $22.85$ & $19.52$ & $29.40$ & $18.52$ & $25.46$ & $24.15$ & $24.70$ \\
Ours                      & $\mathbf{32.70}$ & $\mathbf{25.17}$ & $\mathbf{20.92}$ & $\mathbf{31.16}$ & $\mathbf{20.36}$ & $\mathbf{27.40}$ & $\mathbf{26.80}$ & $\mathbf{27.45}$ \\
\hline 
\multicolumn{9}{c}{}
\end{tabular}
} \vspace{2mm}
\resizebox{\textwidth}{!}{
\begin{tabular}{l|cccccccc}
\multicolumn{9}{c}{SSIM$\uparrow$}\\
& Room & Fern & Leaves & Fortress & Orchids & Flower & T-Rex & Horns \\
\hline
SRN~\cite{srn}            & $0.883$ & $0.611$ & $0.520$ & $0.641$ & $0.449$ & $0.738$ & $0.761$ & $0.742$ \\
LLFF~\cite{mildenhall19}  & $0.932$ & $0.753$ & $\mbf{0.697}$ & $0.872$ & $0.588$ & $\mbf{0.844}$ & $0.857$ & $\mbf{0.840}$ \\
Ours                      & $\mbf{0.948}$ & $\mbf{0.792}$ & $0.690$ & $\mbf{0.881}$ & $\mbf{0.641}$ & $0.827$ & $\mbf{0.880}$ & $0.828$ \\
\hline
\multicolumn{9}{c}{}
\end{tabular}
} \vspace{2mm}
\resizebox{\textwidth}{!}{
\begin{tabular}{l|cccccccc}
\multicolumn{9}{c}{LPIPS$\downarrow$}\\
& Room & Fern & Leaves & Fortress & Orchids & Flower & T-Rex & Horns \\
\hline
SRN~\cite{srn}            & $0.240$ & $0.459$ & $0.440$ & $0.453$ & $0.467$ & $0.288$ & $0.298$ & $0.376$ \\
LLFF~\cite{mildenhall19}  & $\mathbf{0.155}$ & $\mathbf{0.247}$ & $\mathbf{0.216}$ & $0.173$ & $\mathbf{0.313}$ & $\mathbf{0.174}$ & $\mathbf{0.222}$ & $\mathbf{0.193}$ \\
Ours                      & $0.178$ & $0.280$ & $0.316$ & $\mathbf{0.171}$ & $0.321$ & $0.219$ & $0.249$ & $0.268$ \\
\hline
\end{tabular}
} \vspace{2mm}
\caption{Per-scene quantitative results from our real image dataset. The scenes in this dataset are all captured with a forward-facing handheld cellphone.
}
\label{table:suppresults3}  
\end{table}
\setlength{\tabcolsep}{1.4pt}

\setlength{\tabcolsep}{4pt}
\begin{table}[t]
\centering
\resizebox{\textwidth}{!}{
\begin{tabular}{ll|cccccccc}
&& \multicolumn{8}{c}{PSNR$\uparrow$}\\
&& Chair & Drums & Ficus & Hotdog & Lego & Materials & Mic & Ship  \\
\hline
1) & No PE, VD, H              &      $28.44$  &      $23.11$  &      $25.17$  &      $32.24$  &      $26.38$  &      $24.69$  &      $28.16$  &      $25.12$  \\
2) & No Pos. Encoding          &      $30.33$  &      $24.54$  &      $29.32$  &      $33.16$  &      $27.75$  &      $27.79$  &      $30.76$  &      $26.55$  \\
3) & No View Dependence        &      $30.06$  &      $23.41$  &      $25.91$  &      $32.65$  &      $29.93$  &      $24.96$  &      $28.62$  &      $25.72$  \\
4) & No Hierarchical           &      $31.32$  &      $24.55$  &      $29.25$  &      $35.24$  &      $31.42$  &      $29.22$  &      $31.74$  &      $27.73$  \\
\arrayrulecolor{gray}
\hline
\arrayrulecolor{black}
5) & Far Fewer Images          &      $30.92$  &      $22.62$  &      $24.39$  &      $32.77$  &      $27.97$  &      $26.55$  &      $30.47$  &      $26.57$  \\
6) & Fewer Images              &      $32.19$  &      $23.70$  &      $27.45$  &      $34.91$  &      $31.53$  &      $28.54$  &      $32.33$  &      $27.67$  \\
\arrayrulecolor{gray}
\hline
\arrayrulecolor{black}
7) & Fewer Frequencies         &      $32.19$  & $\mbf{25.29}$ & $\mbf{30.73}$ &      $36.06$  &      $30.77$  & $\mbf{29.77}$ &      $31.66$  &      $28.26$  \\
8) & More Frequencies          &      $32.87$  &      $24.65$  &      $29.92$  &      $35.78$  &      $32.50$  &      $29.54$  &      $32.86$  &      $28.34$  \\
\arrayrulecolor{gray}
\hline
\arrayrulecolor{black}
9) & Complete Model            & $\mbf{33.00}$ &      $25.01$  &      $30.13$  & $\mbf{36.18}$ & $\mbf{32.54}$ &      $29.62$  & $\mbf{32.91}$ & $\mbf{28.65}$ \\
\hline
\multicolumn{9}{c}{}
\end{tabular}
} \vspace{2mm}
\resizebox{\textwidth}{!}{
\begin{tabular}{ll|cccccccc}
&& \multicolumn{8}{c}{SSIM$\uparrow$}\\
&& Chair & Drums & Ficus & Hotdog & Lego & Materials & Mic & Ship  \\
\hline
1) & No PE, VD, H              &      $0.919$  &      $0.896$  &      $0.926$  &      $0.955$  &      $0.882$  &      $0.905$  &      $0.955$  &      $0.810$  \\
2) & No Pos. Encoding          &      $0.938$  &      $0.918$  &      $0.953$  &      $0.956$  &      $0.903$  &      $0.933$  &      $0.968$  &      $0.824$  \\
3) & No View Dependence        &      $0.948$  &      $0.906$  &      $0.938$  &      $0.961$  &      $0.947$  &      $0.912$  &      $0.962$  &      $0.828$  \\
4) & No Hierarchical           &      $0.951$  &      $0.914$  &      $0.956$  &      $0.969$  &      $0.951$  &      $0.944$  &      $0.973$  &      $0.844$  \\
\arrayrulecolor{gray}
\hline
\arrayrulecolor{black}
5) & Far Fewer Images          &      $0.956$  &      $0.895$  &      $0.922$  &      $0.966$  &      $0.930$  &      $0.925$  &      $0.972$  &      $0.832$  \\
6) & Fewer Images              &      $0.963$  &      $0.911$  &      $0.948$  &      $0.971$  &      $0.957$  &      $0.941$  &      $0.979$  &      $0.847$  \\
\arrayrulecolor{gray}
\hline
\arrayrulecolor{black}
7) & Fewer Frequencies         &      $0.959$  & $\mbf{0.928}$ & $\mbf{0.965}$ &      $0.972$  &      $0.947$  & $\mbf{0.952}$ &      $0.973$  &      $0.853$  \\
8) & More Frequencies          &      $0.967$  &      $0.921$  &      $0.962$  &      $0.973$  &      $0.961$  &      $0.948$  & $\mbf{0.980}$ &      $0.853$  \\
\arrayrulecolor{gray}
\hline
\arrayrulecolor{black}
9) & Complete Model            & $\mbf{0.967}$ &      $0.925$  &      $0.964$  & $\mbf{0.974}$ & $\mbf{0.961}$ &      $0.949$  &      $0.980$  & $\mbf{0.856}$ \\
\hline
\multicolumn{9}{c}{}
\end{tabular}
} \vspace{2mm}
\resizebox{\textwidth}{!}{
\begin{tabular}{ll|cccccccc}
&& \multicolumn{8}{c}{LPIPS$\downarrow$}\\
&& Chair & Drums & Ficus & Hotdog & Lego & Materials & Mic & Ship  \\
\hline
1) & No PE, VD, H              &      $0.095$  &      $0.168$  &      $0.084$  &      $0.104$  &      $0.178$  &      $0.111$  &      $0.084$  &      $0.261$  \\
2) & No Pos. Encoding          &      $0.076$  &      $0.104$  &      $0.050$  &      $0.124$  &      $0.128$  &      $0.079$  &      $0.041$  &      $0.261$  \\
3) & No View Dependence        &      $0.075$  &      $0.148$  &      $0.113$  &      $0.112$  &      $0.088$  &      $0.102$  &      $0.073$  &      $0.220$  \\
4) & No Hierarchical           &      $0.065$  &      $0.177$  &      $0.056$  &      $0.130$  &      $0.072$  &      $0.080$  &      $0.039$  &      $0.249$  \\
\arrayrulecolor{gray}
\hline
\arrayrulecolor{black}
5) & Far Fewer Images          &      $0.058$  &      $0.173$  &      $0.082$  &      $0.123$  &      $0.081$  &      $0.079$  &      $0.035$  &      $0.229$  \\
6) & Fewer Images              &      $0.051$  &      $0.166$  &      $0.057$  &      $0.121$  &      $0.055$  &      $0.068$  &      $0.029$  &      $0.223$  \\
\arrayrulecolor{gray}
\hline
\arrayrulecolor{black}
7) & Fewer Frequencies         &      $0.055$  &      $0.143$  & $\mbf{0.038}$ & $\mbf{0.087}$ &      $0.071$  & $\mbf{0.060}$ &      $0.029$  &      $0.219$  \\
8) & More Frequencies          &      $0.047$  &      $0.158$  &      $0.045$  &      $0.116$  &      $0.050$  &      $0.064$  & $\mbf{0.027}$ &      $0.261$  \\
\arrayrulecolor{gray}
\hline
\arrayrulecolor{black}
9) & Complete Model            & $\mbf{0.046}$ & $\mbf{0.091}$ &      $0.044$  &      $0.121$  & $\mbf{0.050}$ &      $0.063$  &      $0.028$  & $\mbf{0.206}$  \\
\hline
\multicolumn{9}{c}{}
\end{tabular}
} \vspace{2mm}
\caption{Per-scene quantitative results from our ablation study. The scenes used here are the same as in Table~\ref{table:suppresults2}.
}
\label{table:suppresultsabl} 
\end{table}
\setlength{\tabcolsep}{1.4pt}


\end{document}